\documentclass[10pt,twocolumn,letterpaper]{article}
\usepackage[margin=0.75in]{geometry}
\usepackage[T1]{fontenc}
\usepackage{amsmath,amssymb}
\usepackage{graphicx,booktabs,array}
\usepackage[numbers,sort&compress]{natbib}
\usepackage{xurl}
\usepackage[hidelinks]{hyperref}
\usepackage{tikz,pgfplots}
\pgfplotsset{compat=1.18}
\usetikzlibrary{arrows.meta,positioning}
\usepackage{float,cuted,capt-of}
\usepackage{placeins}
\title{When Experience Helps or Hinders: Sharing and Policy Reuse in Robot Learning}
\author{%
Chongwen Dong\textsuperscript{*}, Mithun Paul Saint-Germain\textsuperscript{$\dagger$}, and Carlo daCunha\textsuperscript{$\ddagger$}\\[0.5em]
{\small\textsuperscript{*}Department of Mechanical Engineering, Northern Arizona University, Flagstaff, AZ, USA}\\
{\small\textsuperscript{$\dagger$}School of Informatics, Computing, and Cyber Systems, Northern Arizona University, Flagstaff, AZ, USA}\\
{\small\textsuperscript{$\ddagger$}Department of Electrical and Computer Engineering, New Jersey Institute of Technology, Newark, NJ, USA}\\[0.3em]
{\small Email: \href{mailto:cd2466@nau.edu}{cd2466@nau.edu}; \href{mailto:carlo.dacunha@njit.edu}{carlo.dacunha@njit.edu}}
}
\date{}
\begin{document}
\twocolumn[\begin{@twocolumnfalse}
\maketitle
\begin{abstract}
Robots can improve through practice, reuse strategies that worked before, and learn from other robots. Combining these sources of experience seems a natural way to improve performance, but can they sometimes work against one another? We investigate this question in simulated navigation, where a language model revises movement strategies from task reports while each robot learns through trial and error. Five connected experiments examine how experience contributes to improvement. We compare all eight combinations of sharing reports, reusing strategies, and retaining controller learning, then use alternative systems to examine what requires strategy revision or learned control. We return to saved stages of learning to separate the roles of historical strategies and receiving controllers, before testing performance under environmental change. Across three matched runs, retaining controller learning reduced mean final cost. However, adding peer reports reversed the average benefit of strategy reuse, and a selected fixed strategy remained competitive. Later in training, consulting history worked better than editing it directly. Peer information changed behavior without consistently improving performance, while the value of a revision depended on the receiving controller and its subsequent learning. Environmental change further showed that high final success could conceal failures during adaptation. Together, these findings show that experience can support improvement or interfere with it, depending on how it shapes a strategy, how the receiving robot executes the change, and what happens as learning continues.
\end{abstract}
\begin{center}\small\textbf{Keywords:} Multi-robot systems; Experience sharing; Policy reuse; Reinforcement learning; Large language models\end{center}
\vspace{1em}
\end{@twocolumnfalse}]
\begingroup
\renewcommand{\thefootnote}{}
\begin{NoHyper}
\footnotetext{Corresponding author: Chongwen Dong (\href{mailto:cd2466@nau.edu}{cd2466@nau.edu}).}
\end{NoHyper}
\endgroup

\section{Introduction}
\label{sec:introduction}
Robots can improve their performance by learning from previous attempts. Through repeated interaction with their environment, they learn which actions are effective and adjust their behavior accordingly~\cite{kober2013rlrobotics}. They can also reuse previously learned policies~\cite{fernandez2006policyreuse} and benefit from experience collected by other robots~\cite{yahya2017collective,gu2017asynchronous}. These approaches offer ways to build on existing knowledge and reduce the effort required to acquire useful behavior. In a system where robots continue learning while exchanging information, however, it becomes important to understand how these sources of experience work together.

Large language models (LLMs) provide new opportunities to use experience in robot learning. They can interpret execution reports, identify possible improvements, and propose changes to a robot's behavior. Recent studies have used LLMs to generate reward functions~\cite{ma2024eureka}, optimize policy parameters from previous execution traces~\cite{benamor2025sas}, and extract lessons that can guide later attempts~\cite{zhao2024expel}. Other methods combine changes to reward functions with continued policy learning~\cite{huang2025roska,li2025lares}. These developments support a process in which robots perform a task, report the outcome, receive revised guidance, and continue learning under that guidance.

The challenge is that a useful suggestion does not necessarily produce a useful change in behavior. Its effect also depends on the robot that receives it. For example, movement settings that work well for one robot may be less suitable for another with a different turning response. A revision may also alter the guidance on which a controller's previous learning was based. The controller may adapt successfully, require additional practice, or perform worse after the change. As a result, providing more experience or preserving more history does not by itself establish that the robot will improve.

A further difficulty is that experience can be retained in different forms. An explicit policy can preserve movement settings and action rules, while a learned controller preserves knowledge about which actions to select. Reports from other robots provide another form of history: information about what was attempted and what happened. These forms of experience can influence one another. When a language model revises a policy using shared reports, the resulting behavior reflects both that revision and the controller's existing learning. Measuring only the final outcome makes it difficult to determine which part of this process helped and which introduced problems.

Our study examines that interaction in a  simulated robot navigation setting. We build on the framework of ~\cite{dong2026schema}, in which each robot uses a policy that specifies movement parameters and recommends actions, together with a controller that learns through trial and error. After each task round, an LLM reads the available reports and revises the policy for the next round. Robots navigate toward individual goals, and their outcomes are summarized to assess group performance.

Within this framework, we ask when sharing reports, reusing policies, and retaining controller learning help or hinder improvement. Section~\ref{sec:problem_statement} develops this aim into three research questions and outlines the experiments used to address them.



Thus the main contributions of this work are:
\begin{enumerate}

   \item A thorough systematic empirical evaluation of how experience sharing, policy inheritance, and controller learning interact in simulated multi-robot navigation. Across eight configurations, we identify when sharing peer reports reverses the benefit of revising a robot’s previous policy.

\item A paired evaluation of how policy history and prior controller learning shape the value of a revision. At saved checkpoints, we show that consulting a previous policy can outperform editing it directly, and that the same revisions can improve navigation for a less-trained controller while worsening it for a more-trained one.
\end{enumerate}

\begin{figure*}[p] \centering \includegraphics[width=\textwidth,height=0.88\textheight,keepaspectratio]{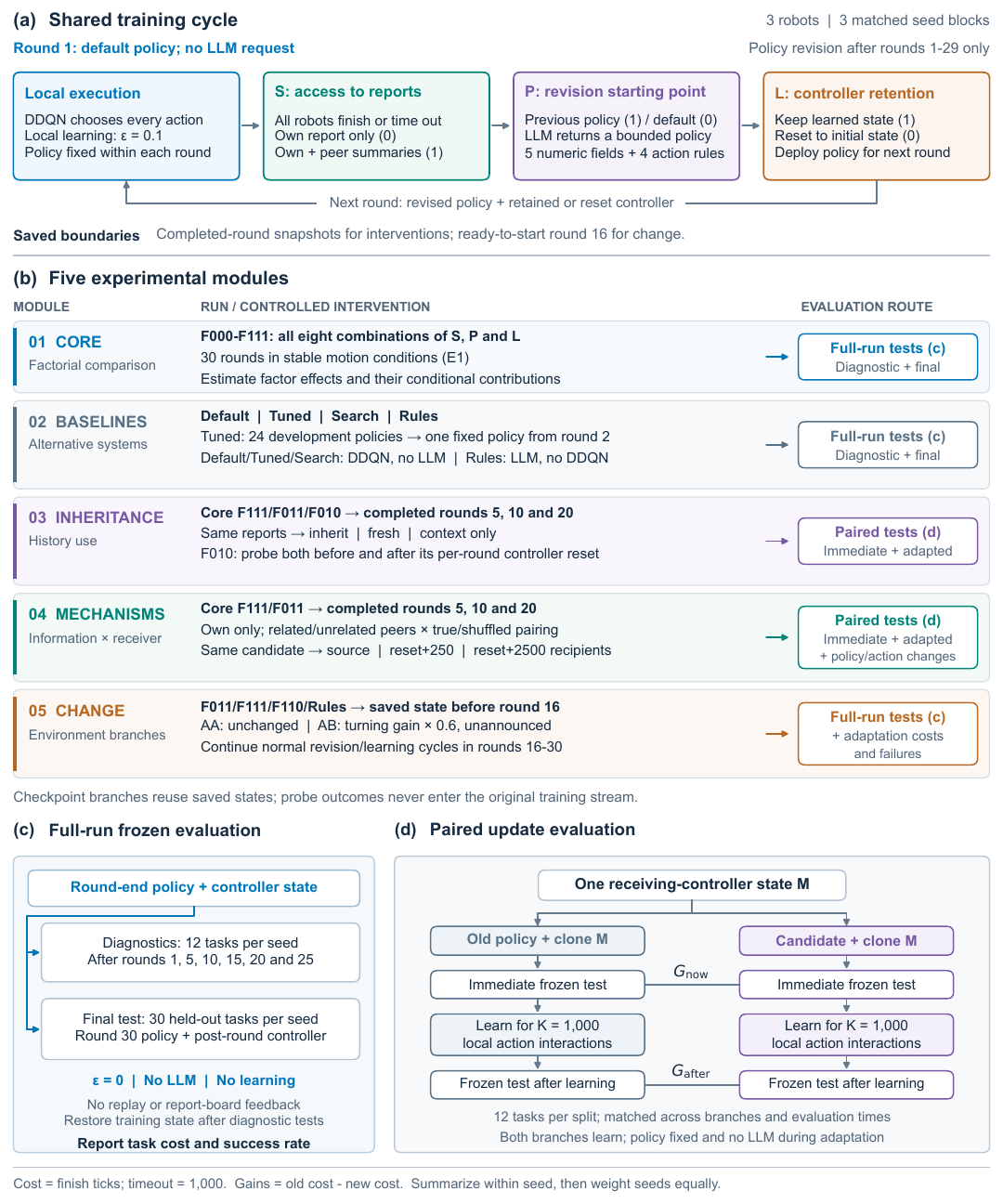} \caption{Experimental and evaluation workflow. (a) Sharing $S$, policy inheritance $P$, and learning retention $L$ act at different stages of the training cycle. (b) Five modules compare complete systems and matched checkpoint interventions. (c) Frozen tests disable exploration, learning, and LLM calls. (d) Old and candidate policies are tested immediately and after both receiver copies complete 1,000 local interactions. Positive gains indicate lower completion cost.} \label{fig:learning_framework} \end{figure*}

\section{Related Work}
\label{sec:related_work}

\noindent\textbf{Language Models for Robot Learning.} Language models have been used to connect task descriptions and execution feedback with changes in robot behavior. Reward-generation methods, including Eureka, Text2Reward, DrEureka, and ELEMENTAL, reduce the effort required to design rewards manually and support robot skill acquisition and transfer~\cite{ma2024eureka,xie2024text2reward,ma2024dreureka,chen2025elemental}. Other approaches use language models to revise control parameters, generate initial policies or planning functions, and guide learning through negotiation between robots~\cite{benamor2025sas,zhu2025lamarl,zhuang2025yolo,godfrey2026marlin}.

ROSKA and LaRes bring reward revision and policy learning into the same improvement process. They coordinate changes to reward functions with continued policy learning, improving learning efficiency while reusing earlier policies or training experience~\cite{huang2025roska,li2025lares}. Together, these approaches establish that feedback can support improvement through changes to rewards, explicit guidance, and learned behavior. Their combination motivates examining how a proposed revision interacts with experience the receiving robot has already retained.

\noindent\textbf{Retaining and Sharing Experience.} Previous experience can be preserved as lessons, examples, or executable skills. Reflexion, ExpeL, and ICAL use information from earlier attempts to improve later decisions~\cite{shinn2023reflexion,zhao2024expel,sarch2024ical}. Voyager develops this idea through a reusable library of executable skills that supports continued exploration and transfer~\cite{wang2024voyager}.

When experience comes from other agents, deciding what to share and how to use it becomes important. Off-agent policy optimization and cautiously optimistic knowledge sharing improve learning by controlling the use of other agents' experience or advice~\cite{chen2024offagent,ba2024cons}. ShareMem distinguishes reusable experience from personal preferences and shows that the benefits of sharing depend on source quality and the experience already available locally~\cite{hu2026sharemem}. These studies demonstrate why additional experience does not necessarily provide additional benefit.

For policy revision, access to history leaves open how that history should be used: an earlier policy may serve as a reference or as the starting point for editing. A further question is whether changes prompted by shared reports translate into better execution.

\noindent\textbf{Policy Revision and Learned Capability.} The usefulness of guidance also depends on the controller responsible for carrying it out. HIRO addresses the changing relationship between high-level actions and learned low-level behavior~\cite{nachum2018hiro}, while Hi Robot connects high-level reasoning with low-level execution~\cite{shi2025hirobot}. LEGION preserves and combines previously learned knowledge to solve subsequent robot tasks~\cite{meng2025legion}. ORCHID examines how planners and controllers adapt together and shows that better planning predictions alone do not necessarily improve task success~\cite{grislain2026orchid}. Collectively, these studies highlight the importance of preserving useful capability and maintaining compatibility between guidance and execution.

These findings motivate examining whether a revision suits the receiver's existing learning and whether its value persists through further practice. Using the framework of~\cite{dong2026schema}, we study this relationship alongside the different ways in which history enters revision.

\section{Problem Statement and Research Questions}
\label{sec:problem_statement}

The core question is when experience improves robot navigation and when its different uses interfere. Our experiments follow a sequence from identifying combined effects to examining why they occur and whether they persist. We first compare all eight combinations of report sharing, policy reuse, and learning retention. Fixed-policy, random-search, and direct-rule alternatives then test which parts of the system account for its performance. At saved checkpoints, we distinguish using history as a reference from editing it directly and test the same revisions with controllers at different stages of learning. Finally, we examine continued performance after the robots' turning response changes.

We assess performance through goal-reaching success and task completion cost. The investigation addresses three research questions:
\begin{enumerate}
    \item \textbf{How do sharing, policy inheritance, and learning retention interact?} We examine whether the contribution of each process changes when the other processes are enabled.
    \item \textbf{How does the use of historical information affect policy revision?} We distinguish consulting a previous policy as a reference from editing it directly, and examine how peer reports influence revisions, action choices, and task outcomes.
    \item \textbf{How does the receiving controller affect the value of a revision?} We examine the same proposed change across controller learning states, compare its immediate effects with those after further learning, and assess performance and failures during adaptation to changed motion conditions.
\end{enumerate}

\section{Methodology}
\label{sec:method}

Figure~\ref{fig:learning_framework} connects the five experimental modules to the learning cycle and their evaluation procedures.

\subsection{Task, policy, and learning cycle}
\label{sec:common_framework}

\noindent\textbf{Navigation task.} Robots navigate from different starting positions toward individual goals in a two-dimensional environment. Each robot observes its distance from the goal and its signed heading error, which indicates the direction and amount of turning needed to face the goal. Robots can exchange reports between rounds, but they do not physically cooperate to complete a shared task. Group performance therefore summarizes individual navigation outcomes.

A trial ends when the robot reaches its goal or exhausts the episode time limit. We measure completion cost in simulation ticks, assigning the time limit to unsuccessful trials. This measure captures the cost of executing a task, rather than language-model latency or computational expense.

The common environment and experimental settings are summarized in Table~\ref{tab:experimental_setup}.

\begin{table*}[!tp]
\centering\small
\caption{Environment and common experimental settings.}
\label{tab:experimental_setup}
\setlength{\tabcolsep}{5pt}
\renewcommand{\arraystretch}{1.15}
\begin{tabular}{@{}>{\raggedright\arraybackslash}p{0.22\textwidth}>{\raggedright\arraybackslash}p{0.74\textwidth}@{}}
\toprule
Setting & Specification \\
\midrule
Independent replication & Three matched runs. \\
Robots and environment & Three robots in an obstacle-free, 101 × 101-patch world. Boundaries do not wrap, and positions are clipped at the world limits. Robots independently navigate toward a common goal. \\
Task generation & Goal coordinates are sampled independently and uniformly from $[-5, 5]$. Each robot’s initial distance from the goal is uniform over [25, 45]. Initial bearing and heading are uniform over a full circle. \\
Trial termination & Arrival occurs when a movement segment intersects the unit square centered on the goal. Trials otherwise terminate at 1000 ticks. \\
Run matching & Task sequences, initial robot states, motion conditions, and controller initialization are matched within each seed block. Model assignments remain fixed within runs and rotate cyclically across blocks. \\
Language-model configuration & Model identities and inference settings are recorded in Table~\ref{tab:llm_specifications}. \\
Controller configuration & Double DQN with a 14 → 64 → 64 → 5 network and ReLU hidden layers; Adam learning rate 0.001; discount factor 0.99; replay capacity 5000; batch size 32; target-network synchronization every 100 gradient updates. Training uses fixed $\epsilon = 0.1$, and policy recommendations receive weight $\beta = 0.05$. \\
Held-out evaluation & Evaluation uses tasks separate from training, with exploration and learning disabled. Controller learning states remain unchanged during evaluation. \\
\bottomrule
\end{tabular}
\end{table*}

\medskip
\noindent\textbf{Run assignments and matching.}
Core configurations and baselines follow the within-run matching summarized in Table~\ref{tab:experimental_setup}. Runs 1--3 use seeds 1000, 10000, and 100000. Their respective (displacement gain, turning gain) pairs are $(1.0,1.0)$, $(0.7,1.0)$, and $(1.0,0.6)$, shared by all three robots within each run. All conditions begin with the default policy, and controller resets do not restart the exploration sequence.

Run 1 uses \texttt{gemma-4-31b-it}, \texttt{gpt-oss-120b}, and \texttt{muse-glimmer} for robot slots 0, 1, and 2, respectively. This assignment rotates one position to the left in each subsequent run. Each run therefore combines one seed, one motion condition, and one model assignment; their effects cannot be separated in this design.

\medskip
\noindent\textbf{Behavioral policy.} Each robot combines an explicit behavioral policy with a learned action-selection controller. The policy determines movement settings and recommends an action for the current situation. The controller uses these recommendations alongside its learned action values.

For robot $i$ in round $r$, the policy is
\begin{equation}
 g_i^r=(\boldsymbol{u}_i^r,\boldsymbol{\rho}_i^r),
 \label{eq:policy}
\end{equation}
where $\boldsymbol{u}$ contains five numerical parameters and $\boldsymbol{\rho}$ contains four action rules (Table~\ref{tab:policy_fields}).

\begin{table*}[t]
\centering\small
\caption{The nine policy fields and their meanings.}
\label{tab:policy_fields}
\setlength{\tabcolsep}{4pt}
\renewcommand{\arraystretch}{1.15}
\begin{tabular}{@{}>{\raggedright\arraybackslash}p{0.29\textwidth}>{\raggedright\arraybackslash}p{0.14\textwidth}>{\raggedright\arraybackslash}p{0.53\textwidth}@{}}
\toprule
Field & Allowed values & Meaning \\
\midrule
\texttt{step\_size} & $0.5$--$3.0$ & Commanded step length for regular forward actions, with or without turning. \\
\texttt{near\_step\_size} & $0.01$--$1.0$ & Commanded step length for small-step actions, with or without turning. \\
\texttt{turn\_angle} & $3^\circ$--$60^\circ$ & Maximum commanded turning angle toward the goal. \\
\texttt{angle\_threshold} & $1^\circ$--$45^\circ$ & Heading-error threshold $\alpha_{\mathrm{align}}$ used to define alignment. \\
\texttt{near\_distance} & $1.0$--$10.0$ & Goal-distance threshold $d_{\mathrm{near}}$ used to define the near region. \\
\midrule
\texttt{far\_misaligned\_action} & One of five actions & Recommended action when far from the goal and misaligned. \\
\texttt{far\_aligned\_action} & One of five actions & Recommended action when far from the goal and aligned. \\
\texttt{near\_misaligned\_action} & One of five actions & Recommended action when near the goal and misaligned. \\
\texttt{near\_aligned\_action} & One of five actions & Recommended action when near the goal and aligned. \\
\bottomrule
\end{tabular}
\end{table*}

A robot is near the goal when $d\leq d_{\mathrm{near}}$, and aligned when $|\theta|\leq\alpha_{\mathrm{align}}$, where $d$ is goal distance and $\theta$ is heading error. Each rule recommends one of five actions: regular forward movement, small forward movement, turning only, or turning combined with either forward step size.

\medskip
\noindent\textbf{Local controller.} A Double DQN controller receives the current observation, policy parameters, and rule recommendation. Outside exploratory moves, it selects
\begin{equation}
 a=\arg\max_{a'\in\mathcal A}
 \left[Q_{\theta_i}(z,a')+\beta h_g(z,a')\right],
 \label{eq:action_selection}
\end{equation}
where $z$ is the controller input, $Q_{\theta_i}$ is the learned action value, and $h_g$ equals one for the action recommended by the policy and zero otherwise. The weight $\beta$ controls the recommendation's influence. The policy therefore guides action selection without requiring the controller to follow its recommendation.

The policy remains fixed within a round while the controller learns from interaction. Its reward is
\begin{equation}
\begin{aligned}
 r_t={}&-0.01-0.02I_{\mathrm{away}}
 -0.05I_{\mathrm{away}}I_{\mathrm{near}}\\
 &+b_t-0.01I_{\mathrm{near,large}}
 +I_{\mathrm{goal}}-I_{\mathrm{timeout}}.
\end{aligned}
\label{eq:reward}
\end{equation}
Each indicator equals one when its condition holds. $I_{\mathrm{away}}$ marks an increase in goal distance, and $I_{\mathrm{near}}$ indicates that the action began within the near-goal region. $I_{\mathrm{near,large}}$ marks a regular forward action, with or without turning, taken from that region without reaching the goal. The final two terms reward goal completion and penalize unsuccessful timeout. The turn-only term is
\[
 b_t=\begin{cases}
 0.005, & \text{corrective turn only},\\
 -0.01, & \text{other turn-only actions},\\
 0, & \text{other actions}.
 \end{cases}
\]
A corrective turn begins outside the alignment threshold and reduces absolute heading error. The reward thus encourages timely arrival while discouraging movement away from the goal, ineffective turning, and unsuccessful large steps near the goal.

The policy also determines the thresholds used in some reward conditions. A revision can consequently change movement, action preference, and aspects of reward shaping together. Our policy-update experiments measure the combined effect of these changes.

\medskip
\noindent\textbf{Language models and inference settings.} The LLMs were accessed using Jetstream2~\cite{hancock2021jetstream2} and AI-VERDE~\cite{mithun2025ai} through an OpenAI-compatible API. Table~\ref{tab:llm_specifications} identifies the language models and the settings used to produce policy revisions. 

\begin{table*}[t]
\centering\small
\caption{Language-model specifications for the reported experiments.}
\label{tab:llm_specifications}
\setlength{\tabcolsep}{5pt}
\renewcommand{\arraystretch}{1.15}
\begin{tabular}{@{}>{\raggedright\arraybackslash}p{0.38\textwidth}>{\raggedright\arraybackslash}p{0.58\textwidth}@{}}
\toprule
Field & Specification \\
\midrule
Exact model identifiers and versions & Identifiers: \texttt{gemma-4-31b-it}, \texttt{gpt-oss-120b}, and \texttt{muse-glimmer}. Server-side revision identifiers are not recorded. \\
Provider, serving platform, and execution dates & Jetstream2 and AI-VERDE (OpenAI-compatible API access). \\
Temperature and other sampling settings & Temperature 0.2 for all models; top-$p$ and other sampling controls are not explicitly supplied. \\
Requested token limits & No explicit input- or output-token limit is requested; provider-side context-window and output limits are not recorded. \\
Reasoning settings, where applicable & No explicit reasoning setting is supplied; provider defaults are not recorded. \\
Decoding seed, if supported & No decoding seed is supplied. Experiment seeds are not passed to the model API. \\
System and user prompt versions and locations & Fixed system instruction and structured user template in the experiment source, under protocol F-SPL-1.0. Source hashes in experiment manifests and request-level prompt hashes track the templates and output schema; no separate version numbers are assigned. \\
Report template and report ordering & Completed-round context, outcome, movement/action counts, and available controller-learning statistics. Own report first, then shared peers in robot-ID order. No earlier-round history or explicit truncation; peer policies appear only in designated sharing diagnostics. \\
Structured-output format or schema & Requested JSON with exactly nine fields: five finite numerical values and four action rules. Values are clipped to declared bounds; rules use five permitted primitives. Validation is local, without an API-enforced output schema. \\
Any model-specific or module-specific settings & Common inference settings across models. Core: one request attempt; later modules: up to three attempts for provider/transport failures, with 10- and 20-second retry delays. \\
Whether LLM weights are updated during experiments & No. LLM weights remain frozen; policy revision uses inference only. \\
\bottomrule
\end{tabular}
\end{table*}

\medskip
\noindent\textbf{Revision between rounds.} After each round, the language model receives the available task reports and a starting policy, together with the navigation objective and fixed revision instructions. It returns a complete policy whose numerical values and action rules are checked against the permitted output format. Failed requests and invalid outputs return the common default policy.

All requests use reports from the same completed round. The revised policies are installed together before the next round begins, separating policy revision between rounds from controller learning within a round. Table~\ref{tab:llm_specifications} provides the report structure, output requirements, and request settings.

\subsection{Comparing ways of using experience}
\label{sec:system_comparisons}

The first experiments ask whether different forms of experience support one another. We independently vary three controls (Table~\ref{tab:experience_controls}).

\begin{table*}[t]
\centering\small
\caption{Independent controls over the use of experience.}
\label{tab:experience_controls}
\setlength{\tabcolsep}{5pt}
\renewcommand{\arraystretch}{1.15}
\begin{tabular}{@{}>{\raggedright\arraybackslash}p{0.18\textwidth}>{\raggedright\arraybackslash}p{0.38\textwidth}>{\raggedright\arraybackslash}p{0.38\textwidth}@{}}
\toprule
Control & Enabled & Disabled \\
\midrule
Sharing, $S$ & Peer reports supplement the robot's own report. & Only the robot's own report is supplied. \\
Policy reuse, $P$ & Revision starts from the previous policy. & Revision starts from the default policy. \\
Learning retention, $L$ & Controller networks, optimizer state, and replay memory persist between rounds. & The controller returns to its initial state each round. \\
\bottomrule
\end{tabular}
\end{table*}

Disabling retention does not remove reinforcement learning: the controller still learns within each round. The comparison measures the combined consequence of preserving its learning state across rounds.

\medskip
\noindent\textbf{Factorial comparison.} The core experiment includes all eight combinations of these controls. We denote a configuration by F$SPL$. For example, F011 uses self-only reports, starts revisions from the previous policy, and retains controller learning. Comparing the complete set reveals whether the effect of one control changes when the others are enabled.

\medskip
\noindent\textbf{Alternative systems.} Four baselines place these comparisons in context.

\emph{Default} keeps the common default policy fixed while the controller continues learning. It tests what local learning can achieve without policy revision.

\emph{Tuned} selects a fixed policy from 24 candidates: the default and 23 randomly sampled settings of its five numerical fields. All four action rules retain their default values. Candidates begin from matched controller states and receive the same learning procedure on separate development tasks. The candidate with the lowest mean completion cost is installed after the common first round and remains fixed while controller learning continues. \emph{Tuned} tests whether a policy selected in advance can support effective navigation through continued DQN learning, without further policy revision.

\emph{Search} revises policies online without a language model. Each robot retains the policy with its best observed completion time normalized by initial goal distance. After a round, it copies that historical best policy and randomly changes one of its nine fields. The next round supplies the score used to update the historical best. Controller learning continues throughout.

\emph{Rules} follows the language-models generated policy's action recommendations directly, without DQN selecting the actions.

These baselines compare different ways of obtaining and executing policies. Core and baseline runs use 12 diagnostic scenarios at intermediate checkpoints and 30 test scenarios after round 30, with three robots in each scenario.

\subsection{Examining individual policy revisions}
\label{sec:receiving_controller}

Whole-system comparisons show what happens over a sequence of revisions, but they cannot explain the effect of any one update. We therefore return to checkpoints saved after rounds 5, 10, and 20 and compare a candidate revision with continued use of the old policy. These experiments address two questions: how does the presentation of experience influence the candidate, and how does the receiving controller influence its value?

\medskip
\noindent\textbf{Policy history as information or starting point.}
\label{sec:history}
A previous policy can help a language model understand earlier choices, but it can also anchor the next revision to those choices. To distinguish these roles, we generate three candidates from the same completed-round reports (Table~\ref{tab:history_modes}). Task instructions, output requirements, and report ordering remain the same across conditions.

\begin{table}[htbp]
\caption{Separating the two roles of policy history.}
\label{tab:history_modes}
\centering\small
\setlength{\tabcolsep}{4pt}
\begin{tabular}{@{}>{\raggedright\arraybackslash}p{0.25\columnwidth}>{\raggedright\arraybackslash}p{0.31\columnwidth}>{\raggedright\arraybackslash}p{0.33\columnwidth}@{}}
\toprule
Condition & Revision starts from & Previous policy available \\
\midrule
Inherit & Previous policy & As starting point \\
Fresh & Default policy & No \\
Context only & Default policy & As reference \\
\bottomrule
\end{tabular}
\end{table}

Comparing context only with fresh examines the value of access to the previous policy. Comparing inherit with context only examines the effect of using that policy as the revision starting point. We repeat these comparisons with shared feedback and self-only feedback, using F111 and F011 checkpoints, respectively. Both source configurations retain controller learning. F010 provides an additional source with self-only feedback, policy inheritance, and a controller reset between rounds. Its candidates are tested before and after the reset; the history results below focus on F111 and F011.

\medskip
\noindent\textbf{Content of shared reports.}
\label{sec:shared_feedback}
The mechanisms experiment keeps the receiver's own report fixed and varies the peer information supplied to the language model. At each checkpoint, we generate five candidates: one from the receiver's report alone and four that add two peers' records. The four shared conditions combine similar or different motion responses with true or shuffled policy--outcome pairings.

Motion response is described by displacement gain, which scales forward movement, and turning gain, which scales rotation. Similar records come from robots with the same gains as the receiver; different records differ in at least one gain. Records are drawn from the same learning round, with starting distance and heading error matched as closely as possible. Gain values guide record selection but are not shown to the language model.

In true pairings, each peer's policy accompanies its own results. Shuffling preserves the policies and starting situations but exchanges the peers' task outcomes, movement summaries, and learning statistics. This changes which results are attributed to each policy. The comparison examines whether the relationship between a reported policy and its outcome influences the revision.

\medskip
\noindent\textbf{Learning state of the receiving controller.} In the mechanisms experiment, each candidate is generated once and tested on three controller states:
\begin{itemize}
 \item \textbf{Source:} the controller saved at the checkpoint.
 \item \textbf{Reset+250:} the initial controller, trained under the old policy for 250 action interactions.
 \item \textbf{Reset+2500:} the initial controller, trained under the old policy for 2,500 action interactions.
\end{itemize}
These preparation periods create different learning states before the candidate is installed. Holding the candidate fixed prevents differences in generated advice from being mixed with differences in the receiving controller.

\medskip
\noindent\textbf{Paired evaluation before and after learning.} In both the inheritance and mechanisms experiments, we create identical controller copies for each receiving state, as shown in Figure~\ref{fig:learning_framework}(d). One continues with the old policy; the other receives the candidate. Both are first evaluated without exploration or learning, using 12 diagnostic tasks and 12 test tasks per receiving robot. Each then continues learning for 1000 interactions under its assigned policy on the same sequence of tasks, after which both are evaluated again on the same held-out tasks. This learning period is separate from controller preparation and testing.

These paired tests leave the source checkpoint unchanged, and diagnostic evaluation results are not returned to the online revision process.

Let $C(g,M)$ denote mean completion cost for policy $g$ and controller state $M$. We define
\begin{align}
 G_{\mathrm{now}}&=C(g_{\mathrm{old}},M)-C(g_{\mathrm{new}},M),\\
 G_{\mathrm{after}}&=C(g_{\mathrm{old}},M'_{\mathrm{old}})
                    -C(g_{\mathrm{new}},M'_{\mathrm{new}}),
 \label{eq:paired_gains}
\end{align}
where the primed states result from further learning under their respective policies. Positive gain indicates lower cost under the candidate.

Considering both distinguishes an initial disruption followed by recovery from a loss that persists over the observed learning budget. Absolute costs and success rates accompany the gains, because a controller with stronger starting performance may have less room to improve.

To connect revision to behavior, we also measure numerical parameter changes scaled by their allowed ranges, the number of changed action rules, and greedy-action disagreement over a common set of distance--angle observations. These measurements describe how a candidate differs from the old policy; the paired task evaluations establish whether those differences are useful.

\subsection{Performance after environmental change}
\label{sec:retained_capability}

A policy developed under one motion response may become unsuitable when that response changes. The final experiment examines this possibility by branching from the same saved policy and controller state. One branch continues under the original conditions, while the other experiences reduced turning effectiveness beginning at round 16. The change is unannounced, so the first policy used after the change still reflects the earlier environment.

Both branches continue their normal policy-revision and controller-learning cycles through round 30. We compare F011, F111, F110, and Rules to examine adaptation under different combinations of shared feedback, learning retention, and action selection.

Both continuations use the diagnostic and final evaluation settings described in Section~\ref{sec:system_comparisons}. We assess both the adaptation process and the resulting capability. Training records capture completion costs and timeouts while the system responds to the change. Final held-out evaluations, conducted without exploration or learning, measure what it can achieve afterward. Reporting both prevents successful final performance from obscuring costly or unreliable behavior during adaptation.

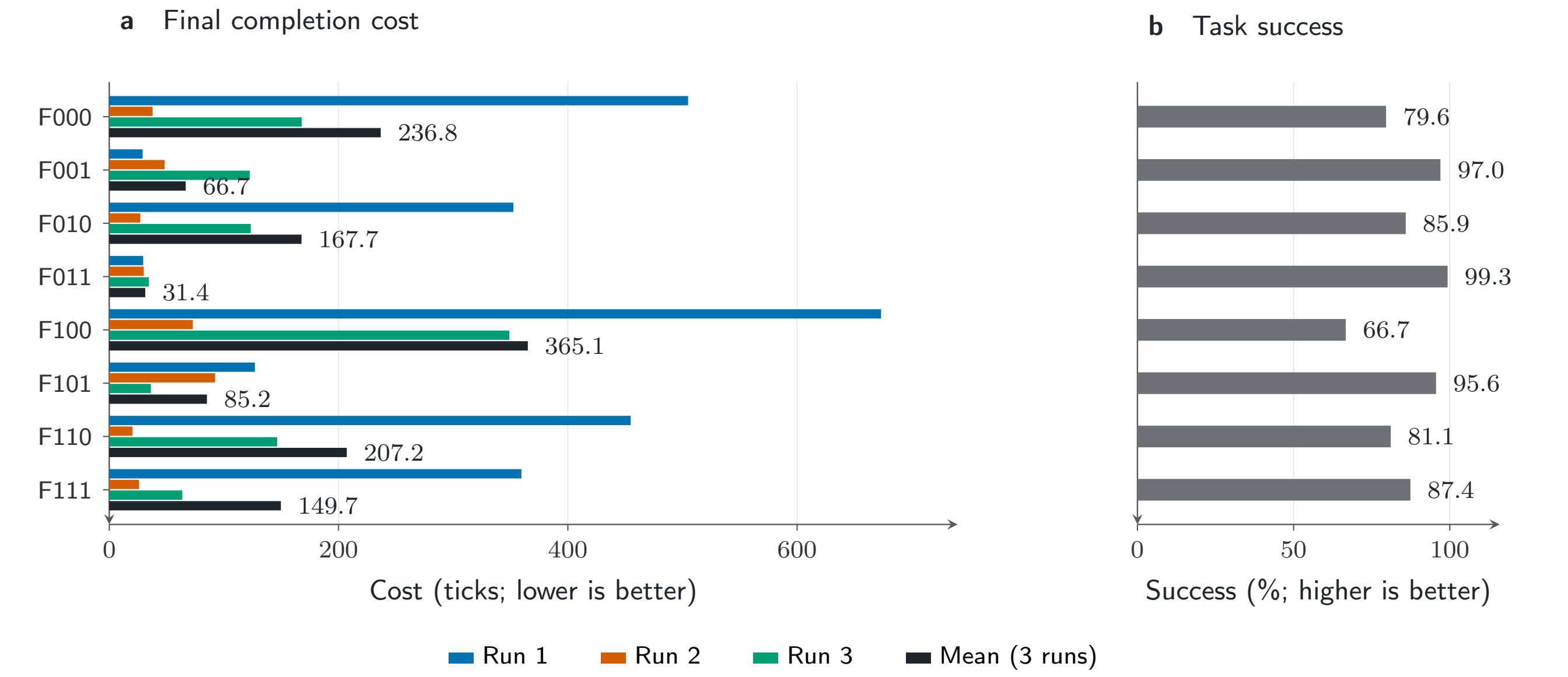
\begin{figure*}[t]
\centering
\begin{minipage}{\textwidth}
\centering
\definecolor{coreone}{HTML}{0072B2}
\definecolor{coretwo}{HTML}{D55E00}
\definecolor{corethree}{HTML}{009E73}
\definecolor{coreink}{HTML}{20242B}
\pgfplotsset{coreaxis/.style={height=6.6cm,width=\linewidth,
axis lines=left,axis line style={black!65,line width=0.45pt},
tick align=outside,tick style={black!65,line width=0.4pt},major tick length=2pt,
xmajorgrids=true,grid style={black!12,line width=0.25pt},scaled x ticks=false,
ymin=-0.65,ymax=7.65,y dir=reverse,ytick={0,1,2,3,4,5,6,7},
tick label style={font=\sffamily\fontsize{8}{9.5}\selectfont,text=black!80},
label style={font=\sffamily\fontsize{9}{10.5}\selectfont,text=coreink},
title style={font=\sffamily\fontsize{9}{11}\selectfont,text=coreink,at={(0,1.04)},anchor=south west},
clip=false}}
\begin{minipage}[t]{0.63\textwidth}
\centering\begin{tikzpicture}
\begin{axis}[coreaxis,title={\textbf{a}\quad Final completion cost},xlabel={Cost (ticks; lower is better)},
xmin=0,xmax=740,xtick={0,200,400,600},yticklabels={F000,F001,F010,F011,F100,F101,F110,F111}]
\addplot[xbar,bar width=3pt,bar shift=0pt,fill=coreone,draw=none] coordinates {(504.9,-0.30) (29.1,0.70) (352.5,1.70) (29.6,2.70) (673.2,3.70) (127,4.70) (454.8,5.70) (359.5,6.70)};
\addplot[xbar,bar width=3pt,bar shift=0pt,fill=coretwo,draw=none] coordinates {(37.8,-0.10) (48.3,0.90) (27.1,1.90) (30.1,2.90) (72.9,3.90) (92.3,4.90) (20.3,5.90) (26,6.90)};
\addplot[xbar,bar width=3pt,bar shift=0pt,fill=corethree,draw=none] coordinates {(167.9,0.10) (122.6,1.10) (123.4,2.10) (34.6,3.10) (349,4.10) (36.3,5.10) (146.4,6.10) (63.7,7.10)};
\addplot[xbar,bar width=3pt,bar shift=0pt,fill=coreink,draw=none,
point meta=x,nodes near coords={\pgfmathprintnumber[fixed,precision=1,fixed zerofill]{\pgfplotspointmeta}},
every node near coord/.append style={anchor=west,xshift=2pt,font=\sffamily\fontsize{8}{9.5}\selectfont,text=coreink}]
coordinates {(236.8,0.30) (66.7,1.30) (167.7,2.30) (31.4,3.30) (365.1,4.30) (85.2,5.30) (207.2,6.30) (149.7,7.30)};
\end{axis}
\end{tikzpicture}
\end{minipage}\hfill
\begin{minipage}[t]{0.32\textwidth}
\centering\begin{tikzpicture}
\begin{axis}[coreaxis,title={\textbf{b}\quad Task success},xlabel={Success (\%; higher is better)},
xmin=0,xmax=116,xtick={0,50,100},yticklabels=\empty,ytick style={draw=none}]
\addplot[xbar,bar width=7pt,fill=coreink!65,draw=none,point meta=x,
nodes near coords={\pgfmathprintnumber[fixed,precision=1,fixed zerofill]{\pgfplotspointmeta}},
every node near coord/.append style={anchor=west,xshift=2pt,font=\sffamily\fontsize{8}{9.5}\selectfont,text=coreink}]
coordinates {(79.6,0.00) (97,1.00) (85.9,2.00) (99.3,3.00) (66.7,4.00) (95.6,5.00) (81.1,6.00) (87.4,7.00)};
\end{axis}
\end{tikzpicture}
\end{minipage}
\par\smallskip
{\sffamily\fontsize{8}{9.5}\selectfont
\tikz[baseline=-0.5ex]{\fill[coreone] (0,-0.06) rectangle (0.28,0.06);}\ Run 1\qquad
\tikz[baseline=-0.5ex]{\fill[coretwo] (0,-0.06) rectangle (0.28,0.06);}\ Run 2\qquad
\tikz[baseline=-0.5ex]{\fill[corethree] (0,-0.06) rectangle (0.28,0.06);}\ Run 3\qquad
\tikz[baseline=-0.5ex]{\fill[coreink] (0,-0.06) rectangle (0.28,0.06);}\ Mean (3 runs)
}
\caption{Final test performance across the eight core combinations. (a) Each group contains three colored bars for individual-run completion costs and a black bar for their equally weighted mean. (b) Bars and labels show mean success rates.}
\label{fig:core_final}
\end{minipage}
\end{figure*}

\section{Evaluation and Results}
\label{sec:results}

We compare corresponding conditions within runs and give each run equal weight after summarizing its task or candidate evaluations.

\subsection{Core: Conditional benefits of combining experience}
\label{sec:results_sharing}

\suppressfloats[t]
\begin{figure}[t]
\centering
\begin{minipage}{\columnwidth}
\centering
\definecolor{runone}{HTML}{0072B2}
\definecolor{runtwo}{HTML}{D55E00}
\definecolor{runthree}{HTML}{009E73}
\definecolor{summaryink}{HTML}{20242B}
\pgfplotsset{
paperaxis/.style={width=\linewidth,height=3.8cm,
axis lines=left,axis line style={black!65,line width=0.45pt},
tick align=outside,tick style={black!65,line width=0.4pt},major tick length=2pt,
ymajorgrids=true,grid style={black!12,line width=0.25pt},scaled y ticks=false,
tick label style={font=\sffamily\fontsize{8}{9.5}\selectfont,text=black!80},
label style={font=\sffamily\fontsize{9}{10.5}\selectfont,text=summaryink},
title style={font=\sffamily\fontsize{9}{11}\selectfont,text=summaryink,at={(0,1.04)},anchor=south west},
clip=false},
paperrunone/.style={runone,line width=0.65pt,solid,mark=o,mark size=2pt,mark options={solid,fill=white,draw=runone,line width=0.7pt}},
paperruntwo/.style={runtwo,line width=0.65pt,dashed,mark=square,mark size=1.9pt,mark options={solid,fill=white,draw=runtwo,line width=0.7pt}},
paperrunthree/.style={runthree,line width=0.65pt,densely dotted,mark=triangle,mark size=2.2pt,mark options={solid,fill=white,draw=runthree,line width=0.7pt}},
papermean/.style={summaryink,line width=1.25pt,mark=diamond*,mark size=2.8pt,mark options={solid,fill=summaryink,draw=white,line width=0.35pt}},
paperzero/.style={black!65,dashed,line width=0.6pt,forget plot}}
\begin{minipage}[t]{0.98\linewidth}
\centering\begin{tikzpicture}
\begin{axis}[paperaxis,title={\textbf{a}\, Learning reset},ylabel={Inheritance gain (ticks)},xmin=-0.18,xmax=1.18,ymin=-270,ymax=270,ytick={-200,-100,0,100,200},xtick={0,1},xticklabels={{\shortstack{Self only\\{\fontsize{7}{8}\selectfont F000 $\to$ F010}}},{\shortstack{Shared\\{\fontsize{7}{8}\selectfont F100 $\to$ F110}}}}]
\addplot[paperzero] coordinates {(-0.18,0) (1.18,0)};
\addplot[paperrunone] coordinates {(0,152.389) (1,218.411)};
\addplot[paperruntwo] coordinates {(0,10.678) (1,52.600)};
\addplot[paperrunthree] coordinates {(0,44.445) (1,202.622)};
\addplot[papermean] coordinates {(0,69.171) (1,157.878)};
\node[font=\sffamily\fontsize{8}{9.5}\selectfont\bfseries,text=summaryink,fill=white,inner sep=1.8pt,anchor=south,yshift=6pt] at (axis cs:0,69.171) {+69.2};
\node[font=\sffamily\fontsize{8}{9.5}\selectfont\bfseries,text=summaryink,fill=white,inner sep=1.8pt,anchor=south,yshift=6pt] at (axis cs:1,157.878) {+157.9};
\end{axis}
\end{tikzpicture}
\end{minipage}\par\medskip
\begin{minipage}[t]{0.98\linewidth}
\centering\begin{tikzpicture}
\begin{axis}[paperaxis,title={\textbf{b}\, Learning retained},xmin=-0.18,xmax=1.18,ymin=-270,ymax=270,ytick={-200,-100,0,100,200},xtick={0,1},xticklabels={{\shortstack{Self only\\{\fontsize{7}{8}\selectfont F001 $\to$ F011}}},{\shortstack{Shared\\{\fontsize{7}{8}\selectfont F101 $\to$ F111}}}}]
\addplot[paperzero] coordinates {(-0.18,0) (1.18,0)};
\addplot[paperrunone] coordinates {(0,-0.467) (1,-232.544)};
\addplot[paperruntwo] coordinates {(0,18.167) (1,66.289)};
\addplot[paperrunthree] coordinates {(0,88.044) (1,-27.389)};
\addplot[papermean] coordinates {(0,35.248) (1,-64.548)};
\node[font=\sffamily\fontsize{8}{9.5}\selectfont\bfseries,text=summaryink,fill=white,inner sep=1.8pt,anchor=south,yshift=6pt] at (axis cs:0,35.248) {+35.2};
\node[font=\sffamily\fontsize{8}{9.5}\selectfont\bfseries,text=summaryink,fill=white,inner sep=1.8pt,anchor=north,yshift=-6pt] at (axis cs:1,-64.548) {$-$64.5};
\end{axis}
\end{tikzpicture}
\end{minipage}
\par\smallskip
{\sffamily\fontsize{8}{9.5}\selectfont
\tikz[baseline=-0.5ex]{\draw[runone,line width=0.65pt] (0,0)--(0.55,0);\draw[runone,fill=white,line width=0.7pt] (0.275,0) circle (1.8pt);}\ Run 1\hspace{7pt}
\tikz[baseline=-0.5ex]{\draw[runtwo,dashed,line width=0.65pt] (0,0)--(0.55,0);\draw[runtwo,fill=white,line width=0.7pt] (0.22,-0.055) rectangle (0.33,0.055);}\ Run 2\\[2pt]
\tikz[baseline=-0.5ex]{\draw[runthree,densely dotted,line width=0.65pt] (0,0)--(0.55,0);\draw[runthree,fill=white,line width=0.7pt] (0.275,0.075)--(0.205,-0.05)--(0.345,-0.05)--cycle;}\ Run 3\hspace{7pt}
\tikz[baseline=-0.5ex]{\draw[summaryink,line width=1.25pt] (0,0)--(0.55,0);\fill[summaryink] (0.275,0.075)--(0.35,0)--(0.275,-0.075)--(0.20,0)--cycle;}\ Mean (3 runs)
}
\caption{Policy inheritance gains under different sharing and learning conditions.}
\label{fig:interactions}
\end{minipage}
\end{figure}
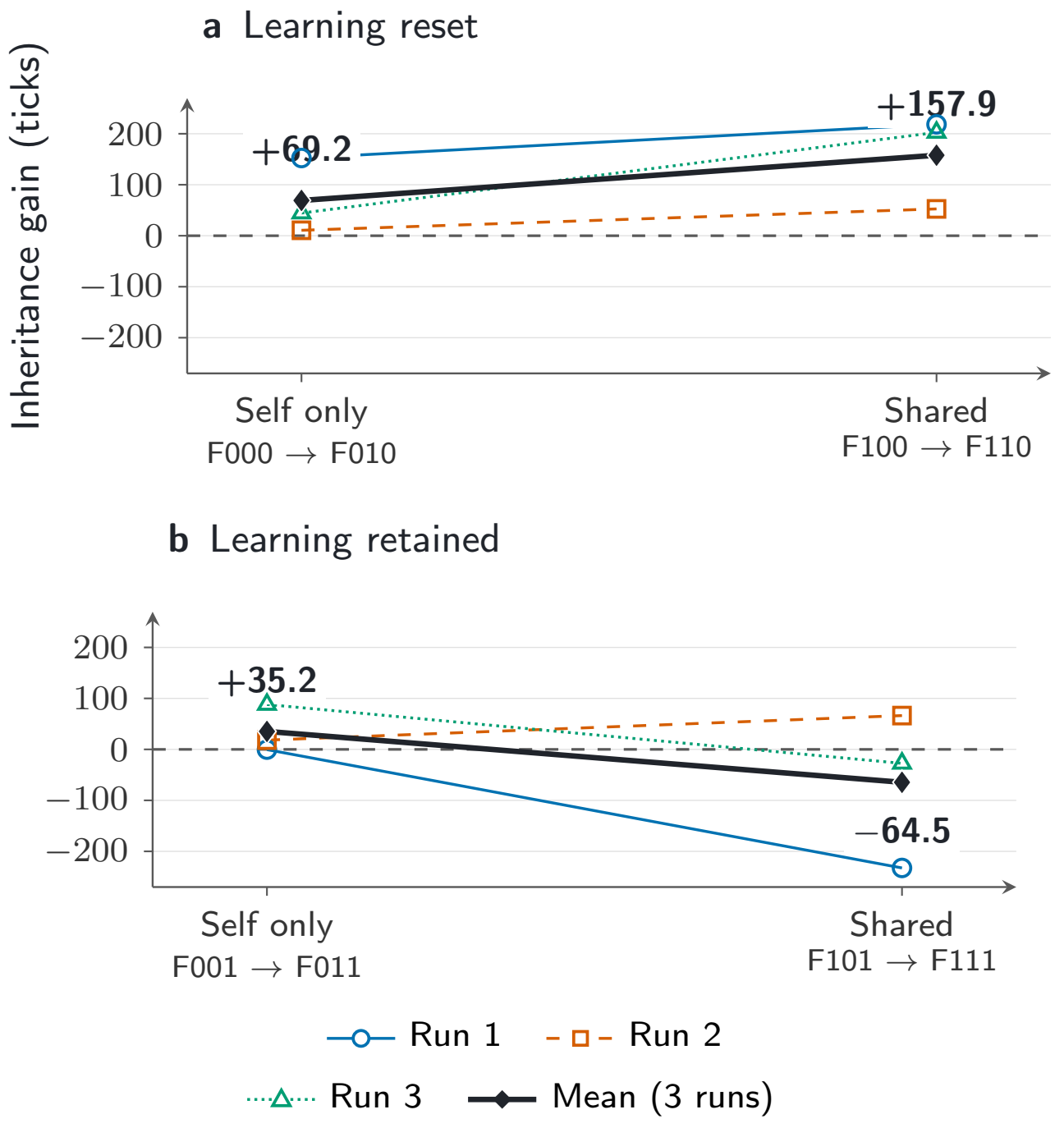

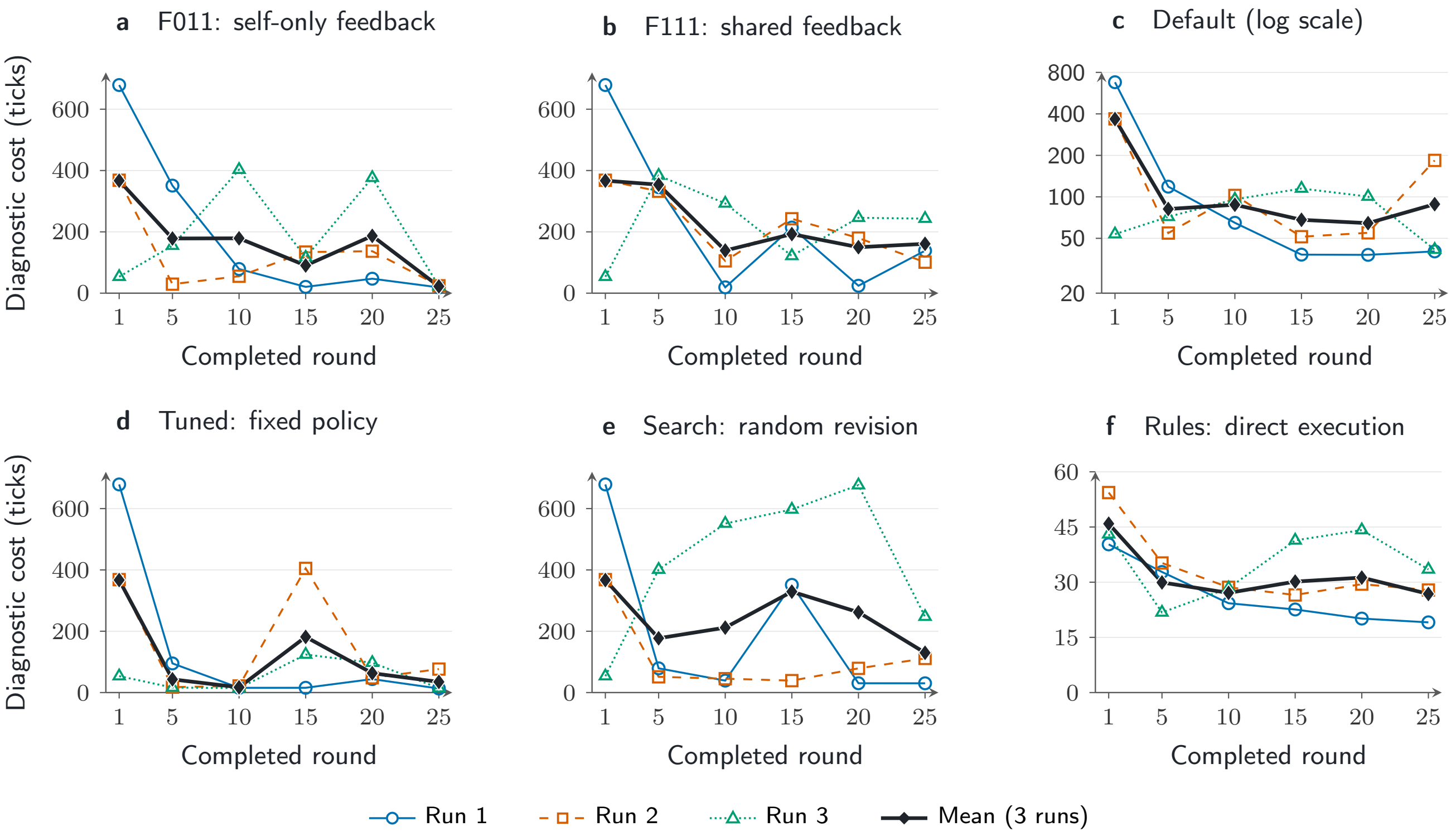
\begin{figure*}[t]
\centering
\begin{minipage}{\textwidth}
\centering
\definecolor{runone}{HTML}{0072B2}
\definecolor{runtwo}{HTML}{D55E00}
\definecolor{runthree}{HTML}{009E73}
\definecolor{summaryink}{HTML}{20242B}
\pgfplotsset{
paperaxis/.style={width=\linewidth,height=4.2cm,
axis lines=left,axis line style={black!65,line width=0.45pt},
tick align=outside,tick style={black!65,line width=0.4pt},major tick length=2pt,
ymajorgrids=true,grid style={black!12,line width=0.25pt},scaled y ticks=false,
tick label style={font=\sffamily\fontsize{8}{9.5}\selectfont,text=black!80},
label style={font=\sffamily\fontsize{9}{10.5}\selectfont,text=summaryink},
title style={font=\sffamily\fontsize{9}{11}\selectfont,text=summaryink,at={(0,1.04)},anchor=south west},
clip=false},
paperrunone/.style={runone,line width=0.65pt,solid,mark=o,mark size=2pt,mark options={solid,fill=white,draw=runone,line width=0.7pt}},
paperruntwo/.style={runtwo,line width=0.65pt,dashed,mark=square,mark size=1.9pt,mark options={solid,fill=white,draw=runtwo,line width=0.7pt}},
paperrunthree/.style={runthree,line width=0.65pt,densely dotted,mark=triangle,mark size=2.2pt,mark options={solid,fill=white,draw=runthree,line width=0.7pt}},
papermean/.style={summaryink,line width=1.25pt,mark=diamond*,mark size=2.8pt,mark options={solid,fill=summaryink,draw=white,line width=0.35pt}},
paperzero/.style={black!65,dashed,line width=0.6pt,forget plot}}
\begin{minipage}[t]{0.32\textwidth}
\centering\begin{tikzpicture}
\begin{axis}[paperaxis,name=trajectory0,title={\textbf{a}\quad F011: self-only feedback},ylabel={Diagnostic cost (ticks)},xlabel={Completed round},xmin=0,xmax=26,ymin=0,ymax=720,ytick={0,200,400,600},xtick={1,5,10,15,20,25}]
\addplot[paperrunone] coordinates {(1,678.806) (5,350.778) (10,78.861) (15,20.583) (20,47.167) (25,18.889)};
\addplot[paperruntwo] coordinates {(1,368.389) (5,29.417) (10,55.306) (15,134.000) (20,136.917) (25,24.139)};
\addplot[paperrunthree] coordinates {(1,53.639) (5,154.889) (10,402.667) (15,116.306) (20,376.500) (25,23.306)};
\addplot[papermean] coordinates {(1,366.944) (5,178.361) (10,178.944) (15,90.296) (20,186.861) (25,22.111)};
\end{axis}
\end{tikzpicture}
\end{minipage}\hfill
\begin{minipage}[t]{0.32\textwidth}
\centering\begin{tikzpicture}
\begin{axis}[paperaxis,name=trajectory1,title={\textbf{b}\quad F111: shared feedback},xlabel={Completed round},xmin=0,xmax=26,ymin=0,ymax=720,ytick={0,200,400,600},xtick={1,5,10,15,20,25}]
\addplot[paperrunone] coordinates {(1,678.806) (5,345.528) (10,19.111) (15,214.056) (20,24.250) (25,138.500)};
\addplot[paperruntwo] coordinates {(1,368.389) (5,332.556) (10,105.306) (15,242.639) (20,179.722) (25,101.111)};
\addplot[paperrunthree] coordinates {(1,53.639) (5,383.528) (10,292.750) (15,121.528) (20,245.917) (25,243.083)};
\addplot[papermean] coordinates {(1,366.944) (5,353.870) (10,139.056) (15,192.741) (20,149.963) (25,160.898)};
\end{axis}
\end{tikzpicture}
\end{minipage}\hfill
\begin{minipage}[t]{0.32\textwidth}
\centering\begin{tikzpicture}
\begin{axis}[paperaxis,name=trajectory2,title={\textbf{c}\quad Default (log scale)},xlabel={Completed round},xmin=0,xmax=26,ymode=log,log basis y=10,ymin=20,ymax=800,ytick={20,50,100,200,400,800},yticklabels={20,50,100,200,400,800},xtick={1,5,10,15,20,25}]
\addplot[paperrunone] coordinates {(1,678.806) (5,118.556) (10,64.833) (15,38.083) (20,37.944) (25,40.194)};
\addplot[paperruntwo] coordinates {(1,368.389) (5,54.611) (10,102.361) (15,51.361) (20,54.833) (25,183.861)};
\addplot[paperrunthree] coordinates {(1,53.639) (5,71.667) (10,95.806) (15,114.917) (20,100.472) (25,41.389)};
\addplot[papermean] coordinates {(1,366.944) (5,81.611) (10,87.667) (15,68.120) (20,64.417) (25,88.481)};
\end{axis}
\end{tikzpicture}
\end{minipage}\par\medskip
\begin{minipage}[t]{0.32\textwidth}
\centering\begin{tikzpicture}
\begin{axis}[paperaxis,name=trajectory3,title={\textbf{d}\quad Tuned: fixed policy},ylabel={Diagnostic cost (ticks)},xlabel={Completed round},xmin=0,xmax=26,ymin=0,ymax=720,ytick={0,200,400,600},xtick={1,5,10,15,20,25}]
\addplot[paperrunone] coordinates {(1,678.806) (5,95.083) (10,15.528) (15,15.611) (20,43.917) (25,13.194)};
\addplot[paperruntwo] coordinates {(1,368.389) (5,17.694) (10,21.472) (15,405.028) (20,48.333) (25,76.556)};
\addplot[paperrunthree] coordinates {(1,53.639) (5,16.472) (10,14.389) (15,123.972) (20,97.278) (25,14.639)};
\addplot[papermean] coordinates {(1,366.944) (5,43.083) (10,17.130) (15,181.537) (20,63.176) (25,34.796)};
\end{axis}
\end{tikzpicture}
\end{minipage}\hfill
\begin{minipage}[t]{0.32\textwidth}
\centering\begin{tikzpicture}
\begin{axis}[paperaxis,name=trajectory4,title={\textbf{e}\quad Search: random revision},xlabel={Completed round},xmin=0,xmax=26,ymin=0,ymax=720,ytick={0,200,400,600},xtick={1,5,10,15,20,25}]
\addplot[paperrunone] coordinates {(1,678.806) (5,79.028) (10,38.972) (15,351.250) (20,30.389) (25,30.278)};
\addplot[paperruntwo] coordinates {(1,368.389) (5,51.306) (10,45.278) (15,38.917) (20,78.917) (25,111.361)};
\addplot[paperrunthree] coordinates {(1,53.639) (5,401.083) (10,551.167) (15,597.167) (20,676.472) (25,247.556)};
\addplot[papermean] coordinates {(1,366.944) (5,177.139) (10,211.806) (15,329.111) (20,261.926) (25,129.731)};
\end{axis}
\end{tikzpicture}
\end{minipage}\hfill
\begin{minipage}[t]{0.32\textwidth}
\centering\begin{tikzpicture}
\begin{axis}[paperaxis,name=trajectory5,title={\textbf{f}\quad Rules: direct execution},xlabel={Completed round},xmin=0,xmax=26,ymin=0,ymax=60,ytick={0,15,30,45,60},xtick={1,5,10,15,20,25}]
\addplot[paperrunone] coordinates {(1,40.278) (5,32.778) (10,24.278) (15,22.583) (20,20.111) (25,19.083)};
\addplot[paperruntwo] coordinates {(1,54.361) (5,35.222) (10,28.611) (15,26.528) (20,29.472) (25,27.861)};
\addplot[paperrunthree] coordinates {(1,43.028) (5,21.778) (10,28.389) (15,41.361) (20,44.167) (25,33.472)};
\addplot[papermean] coordinates {(1,45.889) (5,29.926) (10,27.093) (15,30.157) (20,31.250) (25,26.806)};
\end{axis}
\end{tikzpicture}
\end{minipage}
\par\smallskip
{\sffamily\fontsize{8}{9.5}\selectfont
\tikz[baseline=-0.5ex]{\draw[runone,line width=0.65pt] (0,0)--(0.55,0);\draw[runone,fill=white,line width=0.7pt] (0.275,0) circle (1.8pt);}\ Run 1\qquad
\tikz[baseline=-0.5ex]{\draw[runtwo,dashed,line width=0.65pt] (0,0)--(0.55,0);\draw[runtwo,fill=white,line width=0.7pt] (0.22,-0.055) rectangle (0.33,0.055);}\ Run 2\qquad
\tikz[baseline=-0.5ex]{\draw[runthree,densely dotted,line width=0.65pt] (0,0)--(0.55,0);\draw[runthree,fill=white,line width=0.7pt] (0.275,0.075)--(0.205,-0.05)--(0.345,-0.05)--cycle;}\ Run 3\qquad
\tikz[baseline=-0.5ex]{\draw[summaryink,line width=1.25pt] (0,0)--(0.55,0);\fill[summaryink] (0.275,0.075)--(0.35,0)--(0.275,-0.075)--(0.20,0)--cycle;}\ Mean (3 runs)
}
\caption{Navigation cost during learning in the stable environment.}
\label{fig:trajectories}
\end{minipage}
\end{figure*}

\begin{figure}[t]
\centering
\begin{minipage}{\columnwidth}
\centering
\definecolor{runone}{HTML}{0072B2}
\definecolor{runtwo}{HTML}{D55E00}
\definecolor{runthree}{HTML}{009E73}
\definecolor{summaryink}{HTML}{20242B}
\pgfplotsset{
paperaxis/.style={width=\linewidth,height=5.5cm,
axis lines=left,axis line style={black!65,line width=0.45pt},
tick align=outside,tick style={black!65,line width=0.4pt},major tick length=2pt,
ymajorgrids=true,grid style={black!12,line width=0.25pt},scaled y ticks=false,
tick label style={font=\sffamily\fontsize{8}{9.5}\selectfont,text=black!80},
label style={font=\sffamily\fontsize{9}{10.5}\selectfont,text=summaryink},
title style={font=\sffamily\fontsize{9}{11}\selectfont,text=summaryink,at={(0,1.04)},anchor=south west},
clip=false},
paperrunone/.style={runone,line width=0.65pt,solid,mark=o,mark size=2pt,mark options={solid,fill=white,draw=runone,line width=0.7pt}},
paperruntwo/.style={runtwo,line width=0.65pt,dashed,mark=square,mark size=1.9pt,mark options={solid,fill=white,draw=runtwo,line width=0.7pt}},
paperrunthree/.style={runthree,line width=0.65pt,densely dotted,mark=triangle,mark size=2.2pt,mark options={solid,fill=white,draw=runthree,line width=0.7pt}},
papermean/.style={summaryink,line width=1.25pt,mark=diamond*,mark size=2.8pt,mark options={solid,fill=summaryink,draw=white,line width=0.35pt}},
paperzero/.style={black!65,dashed,line width=0.6pt,forget plot}}
\definecolor{historyinherit}{HTML}{332288}
\definecolor{historyfresh}{HTML}{CC6677}
\definecolor{historycontext}{HTML}{117733}
\begin{minipage}[t]{0.98\linewidth}
\centering\begin{tikzpicture}
\begin{axis}[paperaxis,height=4.1cm,title={\textbf{a}\quad F111: shared feedback},ylabel={Gain after learning (ticks)},xlabel={Source checkpoint},xmin=3,xmax=22,ymin=-180,ymax=150,xtick={5,10,20},ytick={-150,-100,-50,0,50,100,150}]
\addplot[paperzero] coordinates {(3,0) (22,0)};
\addplot[historyinherit,line width=1pt,mark=square*,mark size=2.4pt] coordinates {(5,-39.5) (10,-42.5) (20,-127.1)};
\addplot[historyfresh,line width=1pt,dashed,mark=triangle*,mark size=2.6pt] coordinates {(5,-82.0) (10,21.1) (20,74.3)};
\addplot[historycontext,line width=1pt,mark=*,mark size=2.4pt] coordinates {(5,-53.9) (10,18.1) (20,79.4)};
\end{axis}
\end{tikzpicture}
\end{minipage}\par\medskip
\begin{minipage}[t]{0.98\linewidth}
\centering\begin{tikzpicture}
\begin{axis}[paperaxis,height=4.1cm,title={\textbf{b}\quad F011: self-only feedback},xlabel={Source checkpoint},xmin=3,xmax=22,ymin=-180,ymax=150,xtick={5,10,20},ytick={-150,-100,-50,0,50,100,150}]
\addplot[paperzero] coordinates {(3,0) (22,0)};
\addplot[historyinherit,line width=1pt,mark=square*,mark size=2.4pt] coordinates {(5,114.1) (10,-43.7) (20,-100.3)};
\addplot[historyfresh,line width=1pt,dashed,mark=triangle*,mark size=2.6pt] coordinates {(5,-16.6) (10,-119.6) (20,-8.1)};
\addplot[historycontext,line width=1pt,mark=*,mark size=2.4pt] coordinates {(5,-155.3) (10,2.9) (20,9.1)};
\end{axis}
\end{tikzpicture}
\end{minipage}
\par\smallskip
{\sffamily\fontsize{8}{9.5}\selectfont
\tikz[baseline=-0.5ex]{\draw[historyinherit,line width=1pt] (0,0)--(0.55,0);\fill[historyinherit] (0.22,-0.055) rectangle (0.33,0.055);}\ Inherit\qquad
\tikz[baseline=-0.5ex]{\draw[historyfresh,dashed,line width=1pt] (0,0)--(0.55,0);\fill[historyfresh] (0.275,0.075)--(0.205,-0.05)--(0.345,-0.05)--cycle;}\ Fresh\qquad
\tikz[baseline=-0.5ex]{\draw[historycontext,line width=1pt] (0,0)--(0.55,0);\fill[historycontext] (0.275,0) circle (2pt);}\ Context only
}
\caption{The value of different uses of policy history across learning stages.}
\label{fig:history_results}
\end{minipage}
\end{figure}
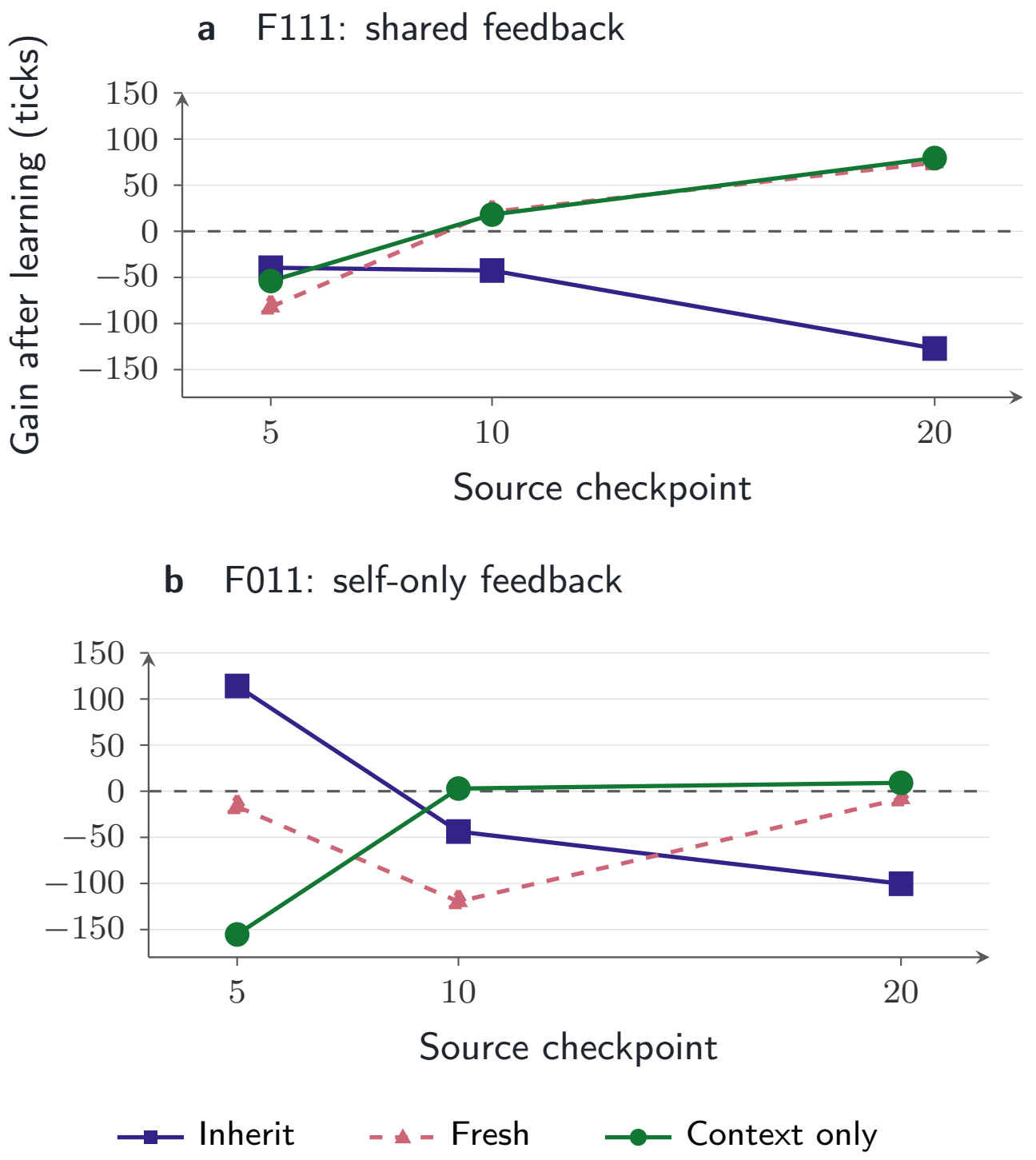

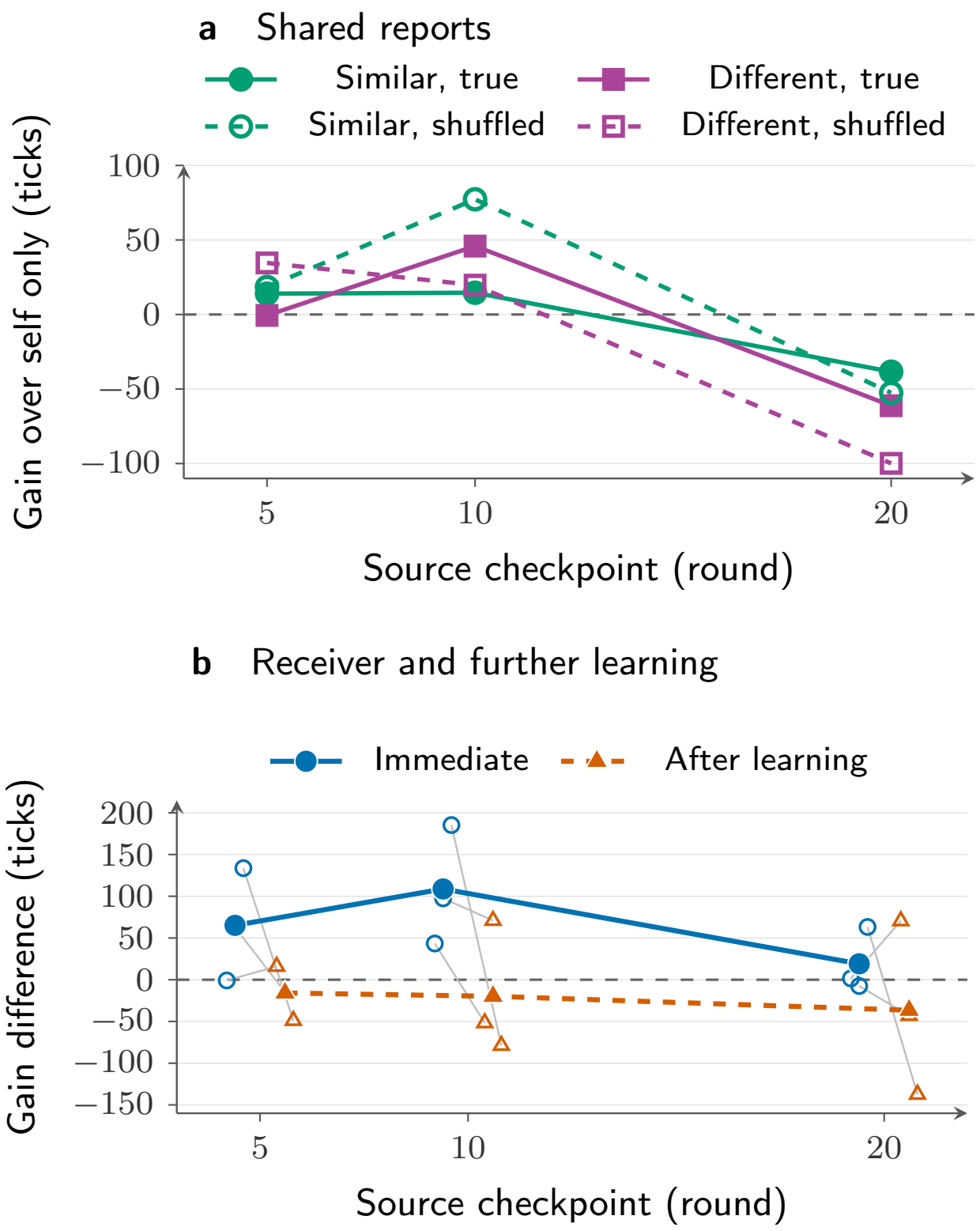
\begin{figure}[t]
\centering
\begin{minipage}{\columnwidth}
\centering
\definecolor{mechsimilar}{HTML}{009E73}
\definecolor{mechdifferent}{HTML}{AA4499}
\definecolor{mechimmediate}{HTML}{0072B2}
\definecolor{mechafter}{HTML}{D55E00}
\pgfplotsset{mechaxis/.style={
width=0.96\linewidth,height=4.2cm,
axis lines=left,axis line style={black!65,line width=0.45pt},
tick align=outside,tick style={black!65,line width=0.4pt},major tick length=2pt,
ymajorgrids=true,grid style={black!12,line width=0.25pt},scaled y ticks=false,
xlabel={Source checkpoint (round)},xtick={5,10,20},
tick label style={font=\sffamily\fontsize{8}{9.5}\selectfont,text=black!80},
label style={font=\sffamily\fontsize{9}{10.5}\selectfont},
title style={font=\sffamily\fontsize{9}{11}\selectfont,at={(0,1.25)},anchor=south west},
legend style={at={(0.5,1.02)},anchor=south,draw=none,font=\sffamily\fontsize{8}{9.5}\selectfont,column sep=5pt},
legend columns=2}}
\begin{minipage}[t]{0.98\linewidth}
\centering
\begin{tikzpicture}
\begin{axis}[mechaxis,title={\textbf{a}\quad Shared reports},
xmin=3,xmax=22,ymin=-110,ymax=100,ytick={-100,-50,0,50,100},ylabel={Gain over self only (ticks)}]
\addplot[black!60,dashed,line width=0.6pt,forget plot] coordinates {(3,0) (22,0)};
\addplot[mechsimilar,line width=1.05pt,mark=*,mark size=2.4pt] coordinates {(5,14.0) (10,14.6) (20,-38.4)};
\addlegendentry{Similar, true}
\addplot[mechdifferent,line width=1.05pt,mark=square*,mark size=2.2pt] coordinates {(5,-0.6) (10,45.9) (20,-61.2)};
\addlegendentry{Different, true}
\addplot[mechsimilar,dashed,line width=1.05pt,mark=o,mark size=2.4pt,mark options={solid,fill=white}] coordinates {(5,18.5) (10,77.3) (20,-52.6)};
\addlegendentry{Similar, shuffled}
\addplot[mechdifferent,dashed,line width=1.05pt,mark=square,mark size=2.2pt,mark options={solid,fill=white}] coordinates {(5,34.6) (10,19.7) (20,-99.9)};
\addlegendentry{Different, shuffled}
\end{axis}
\end{tikzpicture}
\end{minipage}\par\medskip
\begin{minipage}[t]{0.98\linewidth}
\centering
\begin{tikzpicture}
\begin{axis}[mechaxis,title={\textbf{b}\quad Receiver and further learning},
xmin=3,xmax=22,ymin=-160,ymax=215,ytick={-150,-100,-50,0,50,100,150,200},ylabel={Gain difference (ticks)}]
\addplot[black!60,dashed,line width=0.6pt,forget plot] coordinates {(3,0) (22,0)};
\addplot[black!25,line width=0.45pt,forget plot] coordinates {(4.2,-0.88) (5.4,15.93)};
\addplot[black!25,line width=0.45pt,forget plot] coordinates {(4.4,62.61) (5.6,-14.73)};
\addplot[black!25,line width=0.45pt,forget plot] coordinates {(4.6,133.70) (5.8,-48.89)};
\addplot[black!25,line width=0.45pt,forget plot] coordinates {(9.2,43.43) (10.4,-51.71)};
\addplot[black!25,line width=0.45pt,forget plot] coordinates {(9.4,97.48) (10.6,70.95)};
\addplot[black!25,line width=0.45pt,forget plot] coordinates {(9.6,185.30) (10.8,-78.77)};
\addplot[black!25,line width=0.45pt,forget plot] coordinates {(19.2,1.50) (20.4,70.51)};
\addplot[black!25,line width=0.45pt,forget plot] coordinates {(19.4,-7.51) (20.6,-42.93)};
\addplot[black!25,line width=0.45pt,forget plot] coordinates {(19.6,63.24) (20.8,-137.36)};
\addplot[mechimmediate,only marks,mark=o,mark size=1.8pt,mark options={fill=white,line width=0.65pt},forget plot] coordinates {(4.2,-0.88) (4.4,62.61) (4.6,133.70) (9.2,43.43) (9.4,97.48) (9.6,185.30) (19.2,1.50) (19.4,-7.51) (19.6,63.24)};
\addplot[mechafter,only marks,mark=triangle,mark size=2pt,mark options={fill=white,line width=0.65pt},forget plot] coordinates {(5.4,15.93) (5.6,-14.73) (5.8,-48.89) (10.4,-51.71) (10.6,70.95) (10.8,-78.77) (20.4,70.51) (20.6,-42.93) (20.8,-137.36)};
\addplot[mechimmediate,line width=1.2pt,mark=*,mark size=2.7pt,mark options={draw=white,line width=0.35pt}] coordinates {(4.4,65.143) (9.4,108.737) (19.4,19.077)};
\addlegendentry{Immediate}
\addplot[mechafter,dashed,line width=1.2pt,mark=triangle*,mark size=3pt,mark options={solid,draw=white,line width=0.35pt}] coordinates {(5.6,-15.897) (10.6,-19.843) (20.6,-36.593)};
\addlegendentry{After learning}
\end{axis}
\end{tikzpicture}
\end{minipage}
\caption{Policy revision gains on test tasks. (a) Gain after learning relative to self-only feedback, averaged across source configurations and receiver states. (b) Reset+250 gain minus Reset+2500 gain, immediately and after further learning. Colored lines show three-run means; hollow points connected in gray show paired run results in (b).}
\label{fig:controller_results}
\end{minipage}
\end{figure}

\begin{figure*}[t]
\centering
\begin{minipage}{0.88\textwidth}
\centering
\definecolor{runone}{HTML}{0072B2}
\definecolor{runtwo}{HTML}{D55E00}
\definecolor{runthree}{HTML}{009E73}
\definecolor{summaryink}{HTML}{20242B}
\pgfplotsset{
paperaxis/.style={width=\linewidth,height=5.5cm,
axis lines=left,axis line style={black!65,line width=0.45pt},
tick align=outside,tick style={black!65,line width=0.4pt},major tick length=2pt,
ymajorgrids=true,grid style={black!12,line width=0.25pt},scaled y ticks=false,
tick label style={font=\sffamily\fontsize{8}{9.5}\selectfont,text=black!80},
label style={font=\sffamily\fontsize{9}{10.5}\selectfont,text=summaryink},
title style={font=\sffamily\fontsize{9}{11}\selectfont,text=summaryink,at={(0,1.04)},anchor=south west},
clip=false},
paperrunone/.style={runone,line width=0.65pt,solid,mark=o,mark size=2pt,mark options={solid,fill=white,draw=runone,line width=0.7pt}},
paperruntwo/.style={runtwo,line width=0.65pt,dashed,mark=square,mark size=1.9pt,mark options={solid,fill=white,draw=runtwo,line width=0.7pt}},
paperrunthree/.style={runthree,line width=0.65pt,densely dotted,mark=triangle,mark size=2.2pt,mark options={solid,fill=white,draw=runthree,line width=0.7pt}},
papermean/.style={summaryink,line width=1.25pt,mark=diamond*,mark size=2.8pt,mark options={solid,fill=summaryink,draw=white,line width=0.35pt}},
paperzero/.style={black!65,dashed,line width=0.6pt,forget plot}}
\begin{minipage}[t]{0.48\linewidth}
\centering\begin{tikzpicture}
\begin{axis}[paperaxis,height=3.5cm,title={\textbf{a}\quad F011},ylabel={Added cost (ticks)},xlabel={Post-change rounds},xmin=19,xmax=31,ymin=-100,ymax=250,ytick={-100,0,100,200},xtick={20,25,30},xticklabels={16--20,21--25,26--30}]
\addplot[paperzero] coordinates {(19,0) (31,0)};
\addplot[paperrunone] coordinates {(20,10.467) (25,1.533) (30,7.600)};
\addplot[paperruntwo] coordinates {(20,49.067) (25,-0.267) (30,8.733)};
\addplot[paperrunthree] coordinates {(20,191.867) (25,175.133) (30,19.600)};
\addplot[papermean] coordinates {(20,83.800) (25,58.800) (30,11.978)};
\end{axis}
\end{tikzpicture}
\end{minipage}\hfill
\begin{minipage}[t]{0.48\linewidth}
\centering\begin{tikzpicture}
\begin{axis}[paperaxis,height=3.5cm,title={\textbf{b}\quad F111},xlabel={Post-change rounds},xmin=19,xmax=31,ymin=-100,ymax=250,ytick={-100,0,100,200},xtick={20,25,30},xticklabels={16--20,21--25,26--30}]
\addplot[paperzero] coordinates {(19,0) (31,0)};
\addplot[paperrunone] coordinates {(20,78.1) (25,42.7) (30,16.3)};
\addplot[paperruntwo] coordinates {(20,1.6) (25,2.9) (30,59.9)};
\addplot[paperrunthree] coordinates {(20,4.5) (25,176.5) (30,28.5)};
\addplot[papermean] coordinates {(20,28.067) (25,74.033) (30,34.900)};
\end{axis}
\end{tikzpicture}
\end{minipage}\par\smallskip
\begin{minipage}[t]{0.48\linewidth}
\centering\begin{tikzpicture}
\begin{axis}[paperaxis,height=3.5cm,title={\textbf{c}\quad F110},ylabel={Added cost (ticks)},xlabel={Post-change rounds},xmin=19,xmax=31,ymin=-100,ymax=250,ytick={-100,0,100,200},xtick={20,25,30},xticklabels={16--20,21--25,26--30}]
\addplot[paperzero] coordinates {(19,0) (31,0)};
\addplot[paperrunone] coordinates {(20,-2.7) (25,-66.1) (30,62.3)};
\addplot[paperruntwo] coordinates {(20,-1.1) (25,52.7) (30,-60.8)};
\addplot[paperrunthree] coordinates {(20,129.0) (25,218.1) (30,42.1)};
\addplot[papermean] coordinates {(20,41.733) (25,68.233) (30,14.533)};
\end{axis}
\end{tikzpicture}
\end{minipage}\hfill
\begin{minipage}[t]{0.48\linewidth}
\centering\begin{tikzpicture}
\begin{axis}[paperaxis,height=3.5cm,title={\textbf{d}\quad Rules},xlabel={Post-change rounds},xmin=19,xmax=31,ymin=-100,ymax=250,ytick={-100,0,100,200},xtick={20,25,30},xticklabels={16--20,21--25,26--30}]
\addplot[paperzero] coordinates {(19,0) (31,0)};
\addplot[paperrunone] coordinates {(20,205.667) (25,138.400) (30,70.933)};
\addplot[paperruntwo] coordinates {(20,71.133) (25,69.467) (30,57.533)};
\addplot[paperrunthree] coordinates {(20,-52.467) (25,226.400) (30,-59.333)};
\addplot[papermean] coordinates {(20,74.778) (25,144.756) (30,23.044)};
\end{axis}
\end{tikzpicture}
\end{minipage}\par\smallskip
{\sffamily\fontsize{8}{9.5}\selectfont
\tikz[baseline=-0.5ex]{\draw[runone,line width=0.65pt] (0,0)--(0.55,0);\draw[runone,fill=white,line width=0.7pt] (0.275,0) circle (1.8pt);}\ Run 1\qquad
\tikz[baseline=-0.5ex]{\draw[runtwo,dashed,line width=0.65pt] (0,0)--(0.55,0);\draw[runtwo,fill=white,line width=0.7pt] (0.22,-0.055) rectangle (0.33,0.055);}\ Run 2\qquad
\tikz[baseline=-0.5ex]{\draw[runthree,densely dotted,line width=0.65pt] (0,0)--(0.55,0);\draw[runthree,fill=white,line width=0.7pt] (0.275,0.075)--(0.205,-0.05)--(0.345,-0.05)--cycle;}\ Run 3\qquad
\tikz[baseline=-0.5ex]{\draw[summaryink,line width=1.25pt] (0,0)--(0.55,0);\fill[summaryink] (0.275,0.075)--(0.35,0)--(0.275,-0.075)--(0.20,0)--cycle;}\ Mean (3 runs)
}

\caption{Added adaptation cost after a reduction in turning effectiveness.}
\label{fig:change_trajectories}
\end{minipage}
\end{figure*}
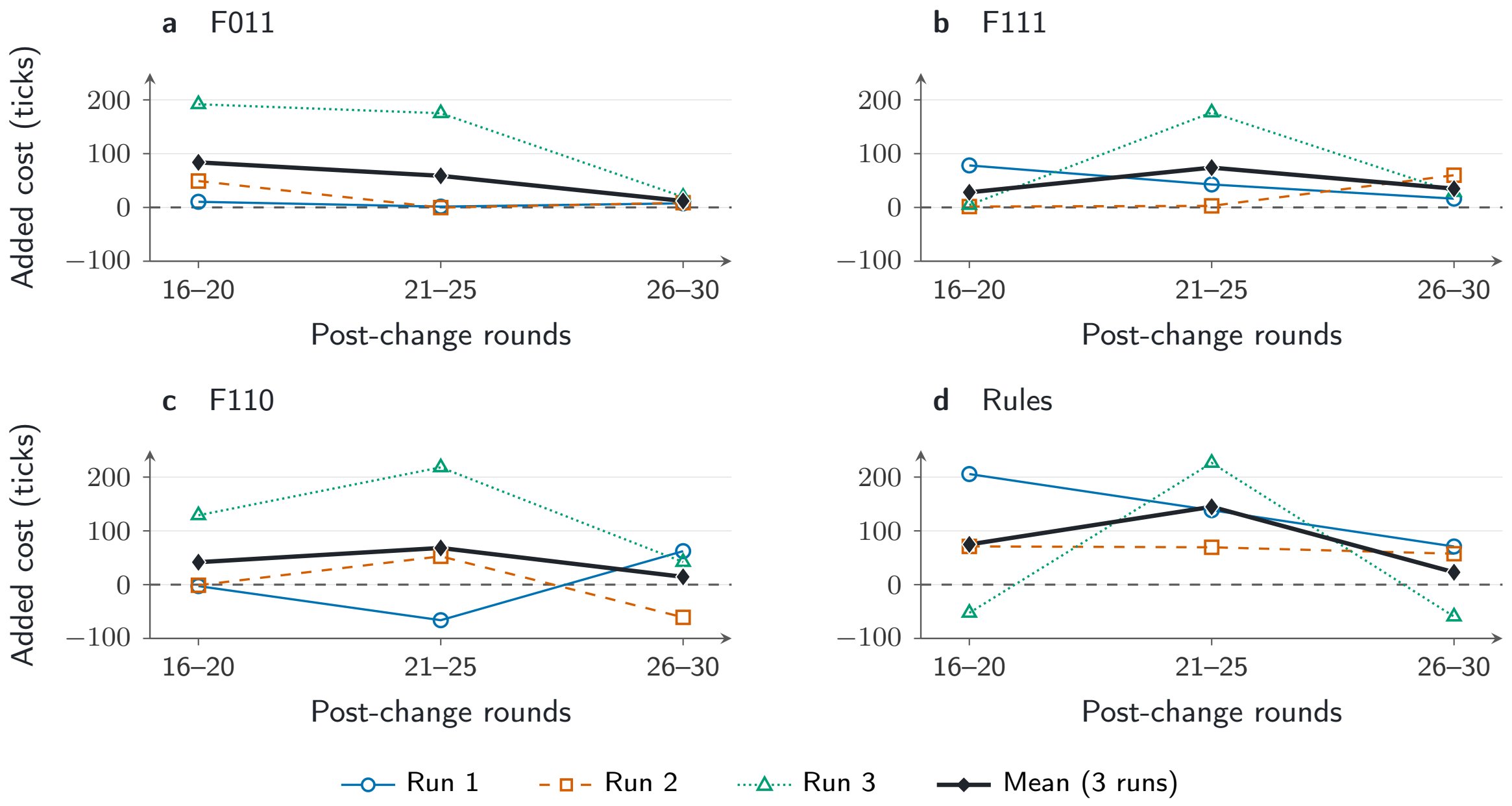

Figure~\ref{fig:core_final} separates the value of accumulating local learning from the value of supplying more information. Retention improved mean final performance under every sharing--inheritance setting, showing the benefit of building on the controller's previous learning. Yet F011, which used self-only reports, outperformed the fully combined F111 configuration in both mean cost and success. Because failed tasks contribute to completion cost, the lower success under F111 shows that its disadvantage involved reliability as well as cost. Adding peer reports did not complement the retained policy and controller learning in this combination; more available experience did not produce more dependable execution.

The inheritance comparison identifies where this failure to combine benefits occurs (Figure~\ref{fig:interactions}). When controllers were reset, inheritance helped on average with or without sharing. Once learning was retained, inheritance helped under self-only feedback (F001 to F011) but hurt under shared feedback (F101 to F111). The same use of policy history thus changed from useful to disruptive depending on the other experience available. Separating the three choices exposes an interaction that the performance of a single complete system would conceal: individual benefits cannot predict the outcome of combining the mechanisms. This reversal describes group means, with variation across runs.

Performance during learning adds another distinction. Retention improved both the final outcome and the mean cost across the later diagnostic checkpoints. Inheritance improved the final average but incurred higher costs across those checkpoints. Its endpoint advantage therefore did not represent a consistently cheaper learning process. Together, these comparisons show that experience sources differ in how they interact and in when their benefits appear. The baseline comparisons next examine whether useful performance requires continuing to revise policies or learning how to execute them.

\subsection{Baselines: What requires policy revision and local learning?}
\label{sec:results_baselines}

The \emph{baselines} show that the value of policy revision depends on what it replaces (Table~\ref{tab:baseline_results}; Figure~\ref{fig:core_final}). F011 had lower final cost than both \emph{Default} and \emph{Search} in every run. Since these references also retained controller learning, the comparisons support a benefit from the tested language-model revisions beyond continued practice with the default policy or online random search. However, \emph{Tuned} performed better than F011 in most runs, while a poorer remaining run raised its overall mean. Therefore a suitably selected fixed policy could  remain competitive, and the best average result did not identify the best approach for every run.

\begin{table}[htbp]
\centering\small\setlength{\tabcolsep}{3pt}
\caption{Final test performance of the four reference systems.}
\label{tab:baseline_results}
\begin{tabular}{@{}lrrrrr@{}}\toprule
System & Run 1 & Run 2 & Run 3 & Mean & Success\\\midrule
Default & 78.4 & 78.9 & 44.8 & 67.4 & 98.1\\
Tuned & 13.8 & 20.9 & 116.5 & 50.4 & 96.7\\
Search & 32.9 & 65.7 & 143.2 & 80.6 & 95.2\\
Rules & 292.1 & 37.6 & 26.2 & 118.6 & 90.7\\
\bottomrule\end{tabular}\end{table}

The learning curves further separate useful components from components that are always necessary (Figure~\ref{fig:trajectories}). \emph{Rules} achieved reliable intermediate performance without a learned action selector, showing that the policy's recommendations could already support effective navigation. Its earlier success did not extend to all final test tasks. \emph{Tuned} also fluctuated and produced failures while its policy remained fixed. These contrasts locate instability on both sides of the revision--execution process: directly following revised rules can lose reliability, while retaining a fixed policy does not ensure stable controller performance.

The whole-system comparisons therefore leave a narrower question: when does a particular revision improve on continued use of the existing policy? We first examine how the previous policy is used to produce that revision.

\subsection{Inheritance: History as information or a starting point}
\label{sec:results_history}

The role of history changed over the course of learning (Figure~\ref{fig:history_results}). Editing the previous policy gave the highest mean gain at the earliest checkpoint, although it still failed to improve on keeping the old policy for the shared-feedback source. At the latest checkpoint, consulting the previous policy as a reference outperformed editing it directly across runs for both sources. Historical information could therefore remain useful even when the historical policy was a poor starting point for revision. The core advantage of inheritance under self-only feedback can thus coexist with harmful inherited updates: a favorable training strategy does not make every revision along that path beneficial.

Fresh clarifies which part of the late improvement came from history. With the shared-feedback source, starting from the default policy without history recovered most of the mean benefit obtained by context only. Much of that benefit was therefore associated with changing the starting point. With the self-only source, context only improved on keeping the old policy while fresh did not, indicating an additional value from consulting history. How history enters revision matters, and the next experiment examines whether the resulting changes help the controller that executes them.

\subsection{Mechanisms: From shared information to execution}
\label{sec:results_controller}

Shared information influenced action choices without reliably improving execution (Figure~\ref{fig:controller_results}(a)). With peers' policies supplied alongside their reports, the shared conditions produced greater action changes but poorer gains than self-only feedback at the latest checkpoint. Earlier benefits did not establish a consistent advantage for correctly paired over shuffled records, or for similar over different motion responses. Thus, neither a stronger response to peer information nor these criteria for selecting relevant peers ensured useful guidance.

The receiver comparison shows why a candidate cannot be judged independently of the controller that executes it. At the middle checkpoint, the mean immediate effect was beneficial for the shorter-trained controller and harmful for the longer-trained one. Yet the longer-trained controller still navigated more efficiently after revision. Update gain and absolute capability therefore answer different questions: a controller may lose more from a revision and still remain the more capable receiver. Existing learning was both a resource for execution and a state that could be disturbed by a new policy, connecting the core interaction to what the receiver had already learned.

Further practice changed the relative value of the updates again (Figure~\ref{fig:controller_results}(b)). The shorter-trained controller improved under the revised policy, but continuing to learn under the old policy produced a better outcome. A before-and-after comparison of the revised branch alone would have credited improvement through practice to the revision. Across checkpoints, the mean gain advantage of the shorter-trained receiver also reversed after further learning, although the direction varied across runs. The paired comparisons therefore explain why immediate gains and larger behavioral changes are insufficient evidence of lasting improvement. The environmental-change experiment examines how this distinction carries over to continued adaptation.

\subsection{Change: Performance under environmental change}
\label{sec:results_sustained}

Environmental change exposed a difference between resisting deterioration and achieving good performance (Figure~\ref{fig:change_trajectories}). F110 could incur less added cost than F011 while still taking longer to complete tasks in the changed environment. A small difference from the unchanged continuation therefore measures sensitivity to change, not navigation quality by itself. The larger adaptation costs across systems in the run that began with weaker turning also show that the burden varied with run conditions. Neither sharing nor retaining controller learning consistently minimized the added cost, so the advantages observed in the stable environment did not translate into uniform protection against change.

Rules illustrates why eventual recovery must also be separated from reliability during adaptation. Its final tests were almost entirely successful despite additional training failures after the change. In one run, its late cost fell below the unchanged continuation after an earlier period of substantial disruption. An evaluation based only on the final test would therefore give an incomplete account of how reliably the system improved.

\section{Discussion}
\label{sec:discussion}

The combined evidence suggests that useful experience can become disruptive when it changes the guidance on which a controller's earlier learning was based. Shared reports and policy history influence what is revised, while retained learning influences how the revision is carried out. The core interaction and checkpoint comparisons therefore describe related parts of the same process: the value of experience depends on how it enters a policy and how the receiver responds. Whether the underlying information is redundant remains unresolved.

For coordinating the three processes, the results favor evaluating an update in relation to the capability already retained by its receiver. Shared content needs an execution-level check, access to history should be separated from direct inheritance, and immediate effects should be assessed alongside effects after further learning. The changed-environment results add a further requirement: final success should be considered together with failures incurred during adaptation and differences across robots. Selective sharing and controller-aware acceptance of updates are practical directions suggested by these comparisons, with their value as online decision rules still to be tested.

The evidence comes from three matched runs in a single navigation setting; each run combines task variation with a motion condition and model assignment. The retention comparison preserves or resets network parameters, optimizer state, and replay memory together, so it cannot separate their individual contributions. Further tasks and controller conditions can establish how widely the observed relationships hold.

\section{Conclusion}
We studied how robots use their own experience and the experience of their peers to revise navigation policies. Across three matched simulation runs, retaining controller learning between rounds reduced mean final navigation cost. Yet the benefits of experience did not simply add up: with learning retained, policy inheritance helped on average under self-only feedback, but adding peer reports reversed that benefit. Comparisons with alternative systems also showed that continued revision was not always necessary for competitive performance. Late in training, consulting an earlier policy as a reference gave better outcomes after further learning than editing it directly. Peer reports could also change behavior without consistently improving execution.

The checkpoint tests further showed that the value of a revision depended on the receiving controller and could change after further learning. Under environmental change, high final success could coexist with failures during adaptation. Together, these findings connect the effects of combining experience to how history shapes a policy, what its receiver has already learned, and how performance develops afterward. Our experiments covered one navigation task in which robots shared reports and moved independently. A next step is to test whether a robot can use what it has already learned, together with a brief trial of a proposed revision, to decide which advice to adopt.

\section*{Acknowledgment}
\textbf{Use of AI tools.} ChatGPT (OpenAI) was used for language assistance,
including grammar correction, wording refinement, and improvements in readability and clarity. All scientific content, methodology, algorithms, code, experiments,
analyses, interpretations, and conclusions were developed and
verified by the authors.


\begin{thebibliography}{99}

\bibitem{kober2013rlrobotics}
J. Kober, J. A. Bagnell, and J. Peters.
\newblock Reinforcement learning in robotics: A survey.
\newblock \emph{The International Journal of Robotics Research}, 32(11):1238--1274, 2013.
\newblock \url{https://doi.org/10.1177/0278364913495721}.

\bibitem{fernandez2006policyreuse}
F. Fern\'andez and M. Veloso.
\newblock Probabilistic Policy Reuse in a Reinforcement Learning Agent.
\newblock In \emph{International Joint Conference on Autonomous Agents and Multiagent Systems (AAMAS)}, 2006.
\newblock Author manuscript: \url{https://www.cs.cmu.edu/afs/cs/user/mmv/www/papers/06aamas-policy-reuse.pdf}.

\bibitem{yahya2017collective}
A. Yahya, A. Li, M. Kalakrishnan, Y. Chebotar, and S. Levine.
\newblock Collective Robot Reinforcement Learning with Distributed Asynchronous Guided Policy Search.
\newblock In \emph{IEEE/RSJ International Conference on Intelligent Robots and Systems (IROS)}, 2017.
\newblock Author manuscript: \url{https://arxiv.org/abs/1610.00673}.

\bibitem{gu2017asynchronous}
S. Gu, E. Holly, T. Lillicrap, and S. Levine.
\newblock Deep Reinforcement Learning for Robotic Manipulation with Asynchronous Off-Policy Updates.
\newblock In \emph{IEEE International Conference on Robotics and Automation (ICRA)}, pages 3389--3396, 2017.
\newblock \url{https://doi.org/10.1109/ICRA.2017.7989385}.

\bibitem{ma2024eureka}
Y. J. Ma, W. Liang, G. Wang, D.-A. Huang, O. Bastani, D. Jayaraman, Y. Zhu, L. Fan, and A. Anandkumar.
\newblock Eureka: Human-Level Reward Design via Coding Large Language Models.
\newblock In \emph{International Conference on Learning Representations (ICLR)}, 2024.
\newblock Author manuscript: \url{https://arxiv.org/abs/2310.12931}.

\bibitem{benamor2025sas}
H. Ben Amor, L. Graesser, A. Iscen, D. D'Ambrosio, S. Abeyruwan,
A. Bewley, Y. Zhou, K. Kalirathinam, S. Mishra, and P. Sanketi.
\newblock SAS-Prompt: Large Language Models as Numerical Optimizers for Robot Self-Improvement.
\newblock In \emph{IEEE International Conference on Robotics and Automation (ICRA)}, 2025.
\newblock Author manuscript: \url{https://arxiv.org/abs/2504.20459v1}.

\bibitem{zhao2024expel}
A. Zhao, D. Huang, Q. Xu, M. Lin, Y.-J. Liu, and G. Huang.
\newblock ExpeL: LLM Agents Are Experiential Learners.
\newblock In \emph{AAAI Conference on Artificial Intelligence}, 38(17):19632--19642, 2024.
\newblock \url{https://doi.org/10.1609/aaai.v38i17.29936}.

\bibitem{huang2025roska}
C. Huang, Y. Chang, J. Lin, J. Liang, R. Zeng, and J. Li.
\newblock Efficient Language-instructed Skill Acquisition via Reward-Policy Co-Evolution.
\newblock In \emph{AAAI Conference on Artificial Intelligence}, 39(14):14576--14584, 2025.
\newblock \url{https://doi.org/10.1609/aaai.v39i14.33597}.

\bibitem{li2025lares}
P. Li, H. Tang, J. Qiao, Y. Zheng, and J. Hao.
\newblock LaRes: Evolutionary Reinforcement Learning with LLM-based Adaptive Reward Search.
\newblock In \emph{Advances in Neural Information Processing Systems}, volume 38, 2025.
\newblock \url{https://doi.org/10.52202/085713-0786}.

\bibitem{dong2026schema}
C. Dong, M. P. Saint-Germain, P. Asif, and C. R. daCunha.
\newblock A Schema Bounded Language Model for Refining Robot Policies Without Destabilizing Local Learning.
\newblock \emph{arXiv preprint arXiv:2609.05133v1}, 2026.
\newblock \url{https://arxiv.org/abs/2609.05133v1}.

\bibitem{xie2024text2reward}
T. Xie, S. Zhao, C. Wu, Y. Liu, Q. Luo, V. Zhong, Y. Yang, and T. Yu.
\newblock Text2Reward: Reward Shaping with Language Models for Reinforcement Learning.
\newblock In \emph{International Conference on Learning Representations (ICLR)}, 2024.
\newblock Author manuscript: \url{https://arxiv.org/abs/2309.11489}.

\bibitem{ma2024dreureka}
Y. J. Ma, W. Liang, H.-J. Wang, Y. Zhu, L. Fan, O. Bastani, and D. Jayaraman.
\newblock DrEureka: Language Model Guided Sim-To-Real Transfer.
\newblock In \emph{Robotics: Science and Systems (RSS)}, 2024.
\newblock \url{https://doi.org/10.15607/RSS.2024.XX.094}.

\bibitem{chen2025elemental}
L. Chen, N. M. Moorman, and M. C. Gombolay.
\newblock ELEMENTAL: Interactive Learning from Demonstrations and Vision-Language Models for Reward Design in Robotics.
\newblock In \emph{International Conference on Machine Learning (ICML)}, PMLR 267:8700--8725, 2025.
\newblock \url{https://proceedings.mlr.press/v267/chen25at.html}.

\bibitem{zhu2025lamarl}
G. Zhu, R. Zhou, W. Ji, and S. Zhao.
\newblock LAMARL: LLM-Aided Multi-Agent Reinforcement Learning for Cooperative Policy Generation.
\newblock \emph{IEEE Robotics and Automation Letters}, 2025.
\newblock Author manuscript: \url{https://arxiv.org/abs/2506.01538v2}.

\bibitem{zhuang2025yolo}
Y. Zhuang, Y. Shen, Z. Zhang, Y. Chen, and F. Miao.
\newblock YOLO-MARL: You Only LLM Once for Multi-Agent Reinforcement Learning.
\newblock In \emph{IEEE/RSJ International Conference on Intelligent Robots and Systems (IROS)}, 2025.
\newblock Author manuscript: \url{https://arxiv.org/abs/2410.03997v2}.

\bibitem{godfrey2026marlin}
T. Godfrey, W. Hunt, and M. D. Soorati.
\newblock MARLIN: Multi-Agent Reinforcement Learning Guided by Language-Based Inter-Robot Negotiation.
\newblock \emph{arXiv preprint arXiv:2410.14383v4}, 2026.
\newblock \url{https://arxiv.org/abs/2410.14383v4}.

\bibitem{shinn2023reflexion}
N. Shinn, F. Cassano, A. Gopinath, K. Narasimhan, and S. Yao.
\newblock Reflexion: Language Agents with Verbal Reinforcement Learning.
\newblock In \emph{Advances in Neural Information Processing Systems}, volume 36, 2023.
\newblock \url{https://doi.org/10.52202/075280-0377}.

\bibitem{sarch2024ical}
G. Sarch, L. Jang, M. J. Tarr, W. W. Cohen, K. Marino, and K. Fragkiadaki.
\newblock VLM Agents Generate Their Own Memories: Distilling Experience into Embodied Programs of Thought.
\newblock In \emph{Advances in Neural Information Processing Systems}, volume 37, 2024.
\newblock \url{https://doi.org/10.52202/079017-2418}.

\bibitem{wang2024voyager}
G. Wang, Y. Xie, Y. Jiang, A. Mandlekar, C. Xiao, Y. Zhu, L. Fan, and A. Anandkumar.
\newblock Voyager: An Open-Ended Embodied Agent with Large Language Models.
\newblock \emph{Transactions on Machine Learning Research}, 2024.
\newblock \url{https://openreview.net/forum?id=ehfRiF0R3a}.

\bibitem{chen2024offagent}
R. Chen, X. Zhang, Y. Du, Y. Zhong, Z. Tian, F. Sun, and Y. Yang.
\newblock Off-Agent Trust Region Policy Optimization.
\newblock In \emph{International Joint Conference on Artificial Intelligence (IJCAI)}, pages 3798--3806, 2024.
\newblock \url{https://doi.org/10.24963/ijcai.2024/420}.

\bibitem{ba2024cons}
Y. Ba, X. Liu, X. Chen, H. Wang, Y. Xu, K. Li, and S. Zhang.
\newblock Cautiously-Optimistic Knowledge Sharing for Cooperative Multi-Agent Reinforcement Learning.
\newblock In \emph{AAAI Conference on Artificial Intelligence}, 38(16):17299--17307, 2024.
\newblock \url{https://doi.org/10.1609/aaai.v38i16.29677}.

\bibitem{hu2026sharemem}
J. Hu, H. Zhao, Q. Jia, D. Chen, T. Zhao, S. Duan, and G. Liu.
\newblock Learning from Others, Acting for You: Cross-User Memory Sharing for LLM Agents.
\newblock \emph{arXiv preprint arXiv:2609.32511v1}, 2026.
\newblock \url{https://arxiv.org/abs/2609.32511v1}.

\bibitem{nachum2018hiro}
O. Nachum, S. Gu, H. Lee, and S. Levine.
\newblock Data-Efficient Hierarchical Reinforcement Learning.
\newblock In \emph{Advances in Neural Information Processing Systems}, volume 31, 2018.
\newblock \url{https://proceedings.neurips.cc/paper/2018/hash/e6384711491713d29bc63fc5eeb5ba4f-Abstract.html}.

\bibitem{shi2025hirobot}
L. X. Shi, B. Ichter, M. R. Equi, L. Ke, K. Pertsch, Q. Vuong, J. Tanner,
A. Walling, H. Wang, N. Fusai, A. Li-Bell, D. Driess, L. Groom, S. Levine, and C. Finn.
\newblock Hi Robot: Open-Ended Instruction Following with Hierarchical Vision-Language-Action Models.
\newblock In \emph{International Conference on Machine Learning (ICML)}, PMLR 267:54919--54933, 2025.
\newblock \url{https://proceedings.mlr.press/v267/shi25d.html}.

\bibitem{meng2025legion}
Y. Meng, Z. Bing, X. Yao, K. Chen, K. Huang, Y. Gao, F. Sun, and A. Knoll.
\newblock Preserving and combining knowledge in robotic lifelong reinforcement learning.
\newblock \emph{Nature Machine Intelligence}, 7:256--269, 2025.
\newblock \url{https://doi.org/10.1038/s42256-025-00983-2}.

\bibitem{grislain2026orchid}
C. Grislain, M. Kappel, O. Sigaud, and M. Chetouani.
\newblock Online Self-Training for Co-Adaptation in Hierarchical Diffusion Policies.
\newblock \emph{arXiv preprint arXiv:2603.05291v2}, 2026.
\newblock \url{https://arxiv.org/abs/2603.05291v2}.


\bibitem{hancock2021jetstream2}
D. Y. Hancock, J. Fischer, J. M. Lowe, W. Snapp-Childs, M. Pierce,
S. Marru, J. E. Coulter, M. Vaughn, B. Beck, N. Merchant, E. Skidmore, and G. Jacobs.
\newblock Jetstream2: Accelerating cloud computing via Jetstream.
\newblock In \emph{Practice and Experience in Advanced Research Computing (PEARC '21)}, Article 11, pp. 1--8. Association for Computing Machinery, 2021.
\newblock \url{https://doi.org/10.1145/3437359.3465565}.

\bibitem{mithun2025ai}
P. Mithun, E. Noriega-Atala, N. Merchant, and E. Skidmore.
\newblock AI-VERDE: A Gateway for Egalitarian Access to Large Language Model-Based Resources For Educational Institutions.
\newblock \emph{arXiv preprint arXiv:2502.09651}, 2025.
\newblock \url{https://doi.org/10.48550/arXiv.2502.09651}.

\end{thebibliography}
\end{document}